%% file: bare_jrnl_compsoc.tex
\documentclass[10pt,journal,compsoc]{IEEEtran}

\ifCLASSOPTIONcompsoc
  \usepackage[nocompress]{cite}
\else
  \usepackage{cite}
\fi

\ifCLASSINFOpdf
  \usepackage[pdftex]{graphicx}
\else
  \usepackage[dvips]{graphicx}
\fi
\usepackage{arydshln} 
\usepackage{multirow}
\usepackage{makecell}
\usepackage{diagbox}
\usepackage{footnote}
\usepackage{algorithm}
\usepackage{algorithmic}
\usepackage{subfigure}
\usepackage{amsmath}
\usepackage{amsfonts}
\usepackage{array}
\usepackage{color}
\usepackage{booktabs}
\usepackage{verbatim}
\usepackage{arydshln}
\usepackage{soul}
\usepackage[dvipsnames]{xcolor}

\newcommand{\zpedit}[1]{{\color{purple} #1}}
\newcommand{\eg}[1]{{\emph{e.g.,} #1}}
\newcommand{\ie}[1]{{{i.e.,} #1}}
\newcommand{\abbr}[1]{{\emph{abbr.} #1}}

\newcommand{\zzcedit}[1]{{\color{black} #1}}
\newcommand{\tabincell}[2]{
\begin{tabular}{@{}#1@{}}#2\end{tabular}
}

\usepackage{url}

\begin{document}
%
\title{TRIE$++$: Towards End-to-End Information Extraction from Visually Rich Documents}

\author{Zhanzhan~Cheng$^*$,
        Peng~Zhang$^*$,
        Can~Li$^*$,
        Qiao~Liang,
        Yunlu~Xu,
        Pengfei~Li,
        Shiliang~Pu,
        Yi~Niu,
        and Fei~Wu
\IEEEcompsocitemizethanks{
\IEEEcompsocthanksitem Z. Cheng and F. Wu are with College of Computer Science and Technology, Zhejiang University, Hangzhou, 310058, China (e-mail: 11821104@zju.edu.cn, wufei@cs.zju.edu.cn). Z. Cheng is also with Hikvision Research Institute, Hangzhou, 310051, China.\protect\\
\IEEEcompsocthanksitem P. Zhang, C. Li, L. Qiao, Y. Xu, P. Li, S. Pu and Y. Niu are with Hikvision Research Institute, Hangzhou, 310051, China (email: zhangpeng23@hikvision.com, lican9@hikvision.com, qiaoliang6@hikvision.com, xuyunlu@hikvision.com, lipengfei27@hikvision.com, pushiliang.hri@hikvision.com, niuyi@hikvision.com).
}

\thanks{$^*$Z. Cheng, P. Zhang and C. Li contributed equally to this research. 
}
}

\markboth{~}%
{Cheng \MakeLowercase{\textit{et al.}}: Bare Demo of IEEEtran.cls for Computer Society Journals}
%


\IEEEtitleabstractindextext{%

\input{sections/0_abstract}

}

\maketitle

\IEEEdisplaynontitleabstractindextext

%
\IEEEpeerreviewmaketitle

\input{sections/1_introduction}

\input{sections/2_related}

\input{sections/3_approach}
\input{sections/4_benchmarks}

\input{sections/5_experiments}
\input{sections/6_conclusion}

\ifCLASSOPTIONcaptionsoff
  \newpage
\fi


\bibliographystyle{IEEEtran}
\bibliography{IEEEabrv,egbib}
\end{document}

%% file: sections/0_abstract.tex
\begin{abstract}
Recently, automatically \zzcedit{extracting information from} visually rich documents (\eg tickets and resumes) has become a hot and vital research topic due to its widespread commercial value. Most existing methods divide this task into two  subparts: the {text reading} part for obtaining the plain text from the original document images and the  {information extraction} part for extracting key contents. 
These methods mainly focus on improving the second, while neglecting that the two parts are highly correlated. 
This paper proposes a unified \zzcedit{end-to-end information extraction framework from visually rich documents}, where  {text reading}  and {information extraction}  can reinforce each other via a well-designed multi-modal context block. 
Specifically, the {text reading} part provides multi-modal features like visual, textual and layout features. 
The multi-modal context block is developed to fuse the generated multi-modal features and even the prior 
knowledge from the pre-trained language model for better semantic representation.  
The {information extraction} part is responsible for generating {key contents} with the fused context  features. 
The framework can be trained in an end-to-end trainable manner, achieving global optimization.
What is more, we define and group visually rich documents into {four} categories across two dimensions, the layout and text type.
For each document category, we provide or recommend the corresponding benchmarks, experimental settings and strong baselines for remedying the problem that this research area lacks the uniform evaluation standard.
Extensive experiments on {four} kinds of benchmarks (from fixed layout to variable layout, from full-structured text to semi-unstructured text) are reported, demonstrating the proposed method's effectiveness. 
Data, source code and models are available. 
\end{abstract}

%% file: sections/1_introduction.tex
\IEEEraisesectionheading{\section{Introduction}\label{sec:introduction}}
\zzcedit{\IEEEPARstart{E}xtracting information from} visually rich document (\abbr VRD) is a traditional yet very important research topic \cite{zhang2020trie,katti2018chargrid,zhao2019cutie,palm2017cloudscan,sage2019recurrent,Aslan2016APB,Janssen2012Receipts2GoTB,dengel2002smartfix,schuster2013intellix,Simon1997AFA}. This is because automatically understanding VRDs can greatly facilitate the key information entry, retrieval and compliance check in enormous and various applications,
including
file understanding in court trial, contract checking in the business system, statements analysis in accounting or financial, case recognition in medical applications, invoice recognition in reimburses system, resume recognition in recruitment system, and automatically examining test paper in education applications, etc.

In general, a VRD \zzcedit{information extraction} system can be divided into two separated parts: text reading and key information extraction. 
Text reading module refers to obtaining text positions as well as their character sequence in document images, which falls into the  computer vision areas related to optical character recognition (\textit{abbr}. OCR)  \cite{wang2020all,qiao2020textperceptron,feng2019textdragon,liao2017textboxes,jaderberg2016reading,wang2012end,shi2016end,liao2019mask}.
Information extraction (\abbr IE) module is responsible for mining key contents (\eg entity, relation) from the captured plain text, related to natural language processing (\abbr NLP) techniques like named entity recognition (\abbr NER)  \cite{nadeau2007survey,lample2016neural,ma2019end} and question-answer  \cite{yang2016stacked,anderson2018bottom,fukui2016multimodal}.
\begin{figure}[t!]
		\centering
		\includegraphics[width=0.5\textwidth]{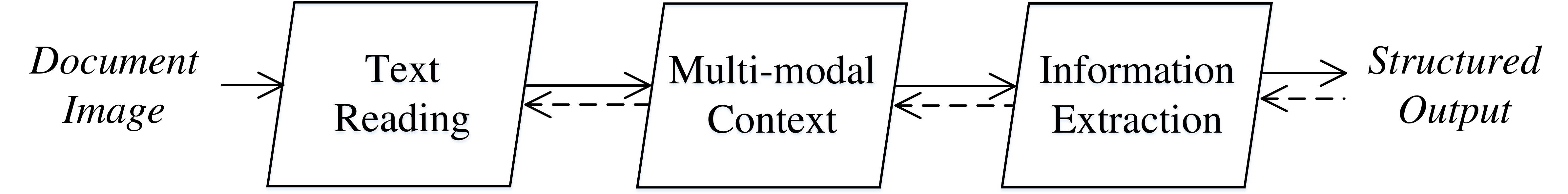}
		\caption{Illustration of the proposed end-to-end VRD \zzcedit{information extraction} framework. It consists of three sub-modules: the text reading part for generating  text layout and character strings, and the information extraction module for outputting key contents. The multi-modal context block is responsible for fully assembling visual, textual, layout features, and even language knowledge, and bridges the text reading and information extraction parts in an end-to-end trainable manner. Dashed lines denote back-propagation. 
	}
		\label{fig:framework}
\end{figure}

Early works ~\cite{palm2017cloudscan,sage2019recurrent}   
implement the VRD \zzcedit{information extraction} frameworks by directly concatenating an offline OCR  engine and the downstream NER-based IE module, which completely discards the visual features and position/layout\footnote{
Note that, terms of `position' and `layout' are two different but highly relevant concepts. 
The former refers to the specific coordinate locations of candidate text regions generated by text reading module. 
The later means the abstract spatial information (\eg position arrangement of text regions) derived from the generated position results via some embedding operations. 
Thus, layout can be treated as the high-level of spatial information in document understanding. 
In the follow-up, we use term `layout' instead of term `position' as one kind of modality.
}
information from images. 
However, as appearing in many applications \cite{palm2017cloudscan,zhang2020trie,dengel2002smartfix,schuster2013intellix,sun2021spatial,wang2021tag}, VRDs are usually organized with both semantic text features and flexible visual structure features in a regular way.
For better results, researchers should consider the key characteristics of documents into their techniques, such as layout, tabular structure, or even font size in addition to the plain text. 
Then recent works begin to incorporate these characteristics into the IE module by embedding multi-dimensional information such as text content and their layouts 
\cite{katti2018chargrid,denk2019bertgrid,zhao2019cutie,palm2019attend,liu2019graph},
 and even image features  \cite{xu2019layoutlm,PICK2020YU,Xu2020LayoutLMv2MP}.

Unfortunately, all existing methods suffer from two main problems: 
First, multi-modality features (like visual, textual and even layout features) are essential for VRD \zzcedit{information extraction}, but the exploitation of the multi-modal features is limited in previous methods. 
Contributions of different kinds of features should be addressed for the IE part. 
For another, text reading and IE modules are highly correlated, but their contribution and relations have rarely been explored. 
\zzcedit{
Therein, the real bottleneck of the whole framework has not been  addressed well. 

Intuitively, it is a way to combine an OCR engine (\eg Tesseract  \cite{smith2007overview}) with a general IE module into a end-to-end framework, while it has several practical issues. 
(1) It is irresponsible to apply the general OCR engines on the specific scenes directly due to the domain gap problems, leading to the dramatic accuracy drop. 
(2) A general and big IE model like LayoutLM \cite{xu2019layoutlm,Xu2020LayoutLMv2MP} may help achieve better IE performance, which however brings extra computational cost.
(3) The non-trainable pipeline strategies will bring extra model maintenance costs and sub-optimal problems.
}

\begin{table*}[th!]
	\caption{Categories of visually rich document scenarios \zzcedit{for information extraction}. A majority of them have been studied in existing research works. 
}
	\label{table:dataset_summary}
    \centering
	\begin{tabular}{c|l|l}
		\diagbox{Layout}{\makecell[c]{Text\\type}} & \multicolumn{1}{c|}{Structured} & \multicolumn{1}{c}{Semi-structured}  \\ 
\cline{1-3}
		Fixed    & \makecell[l]{\textit{Category I}:\\
										value-added tax invoice \cite{liu2019graph}, passport \cite{qin2019eaten}, fixed-format taxi invoice \cite{zhang2020trie},\\
										national ID card \cite{Zhenlong2019TowardsPE}, 
										train ticket \cite{PICK2020YU,Janssen2012Receipts2GoTB}, business license \cite{wang2021tag}} 
				   & \makecell[l]{\textit{Category II}:\\ 
										business email\cite{Harley2015EvaluationOD}, \\
										national housing contract} \\
\cline{1-3}
		Variable & \makecell[l]{\textit{Category III}:\\ 
										medical invoice \cite{PICK2020YU,dengel2002smartfix}, paper head\cite{wang2021towards}, bank card \cite{Zhenlong2019TowardsPE}, \\
										free-format invoice\cite{palm2017cloudscan,katti2018chargrid,palm2019attend,MajumderPTWZN20,Rusiol2013FieldEF,Ha2018RecognitionOO,zhang2020trie}, business card \cite{qin2019eaten},\\
										purchase receipt \cite{zhao2019cutie,liu2019graph,PICK2020YU,Janssen2012Receipts2GoTB,sun2021spatial},
										 purchase orders\cite{sage2019recurrent,xu2019layoutlm}}  
					& \makecell[l]{\textit{Category IV}:\\
										personal resume \cite{zhang2020trie}, \\ 
										financial report \cite{Harley2015EvaluationOD},newspaper\cite{Yang2017LearningTE},\\
										free-format sales contract\cite{Gralinski2020KleisterAN}}     
	\end{tabular}
\end{table*}
Considering the above issues, in this paper, we propose a novel end-to-end \zzcedit{information extraction framework from VRDs, named as TRIE++}. The workflow is as shown in Figure~\ref{fig:framework}. 
Instead of focusing on information extraction task only, we bridge \textit{text reading} and \textit{information extraction} tasks via a developed multi-modal context block.  
In this way, two separated tasks can reinforce each other amidst a unified framework.  
Specifically, 
the text reading module produces diversiform features, including layout features, visual  features and textual features.  
The multi-modal context block fuses multi-modal features with the following steps:
(1) Layout features, visual features and textual features are first fed into the multi-modal embedding module, obtaining their embedding representation. 
(2) Considering the effectiveness of the language model like BERT \cite{denk2019bertgrid}, 
\zzcedit{a prior language knowledge absorption mechanism is developed to incorporate the robust semantic representations of pre-trained language model.}
(3) The embedded features are then correlated with the spatial-aware attention to learn the instance-level interactions. It means  different text instances may have  explicit or implicit interactions, \eg the `Total-Key' and `Total-Value' in receipts are highly correlated.

Consequently, the multi-modal context block can provide robust features for the information extraction module, and the supervisions in information extraction also contribute to the optimization of text reading. 
Since all the modules in the network are differentiable, the whole network could be trained in a global optimization way.
To the best of our knowledge, this is the first end-to-end trainable framework. 
\zzcedit{
To implement it, it is also challenging when considering the fusion of multi-modal information, the global optimization or even prior language knowledge absorption into a framework simultaneously, yet working well on different kinds of documents.
}

We also notice that it is difficult to compare existing methods directly due to the different benchmarks used (most of them are private), the non-uniform evaluation protocols, and even various experimental settings. 
As is known to all, text reading \cite{Chen2020TextRI} is a rapidly growing research area, attributing to its various applications and its uniform benchmarks and evaluation protocols. 
We here reckon that these factors may restrict the study of document understanding. 
To remedy this problem, we first analyze many kinds of documents, and then categorize VRDs into four groups along the dimensions of \emph{layout} and \emph{text type}. 
\textit{Layout} refers to the relative position distribution of texts or text blocks, which contains two modes: the fixed mode and the variable mode.
The former connotes documents that follow a uniform layout format, such as passport and the national value-added tax invoice, while the latter means that documents may appear in different layouts. 
Referring to \cite{judd2004apparatus,soderland1999learning}, we define \textit{text type} into two modalities\footnote{Another text type, the unstructured, is also defined in \cite{judd2004apparatus}, which means that document content is grammatically free text without explicit identifiers such as books. 
Since such documents usually lack visually rich elements (\eg layout), we exclude it from the concept of VRD.}
: 
the structured and the semi-structured. 
In detail, the structured type means that document information is organized in a predetermined schema, i.e., the key-value schema of the document is predefined and often tabular in style, which delimits entities to be extracted directly. 
For example, taxi invoices usually have quite a uniform tabular-like layout and information structure like  `Invoice Number', `Total', `Date' etc. 
The semi-structured type connotes that document content is usually ungrammatical, but each portion of the content is not necessarily organized in a predetermined format. For example, a resume may include some predefined fields such as job experience and education information. Within the job experience fields, the document may include free text to describe the person's job experience. Then, the user may desire to search on free text only within the job experience field. 
Table~\ref{table:dataset_summary} summarizes the categories of visually rich documents from the previous research literature. 
Secondly, we recommend or provide the corresponding benchmarks for each kind of documents, and also provide the uniform evaluation protocols,  experimental settings and strong baselines, expecting to promote this research area. 


Major contributions are summarized as follows. 
(1) We propose an end-to-end trainable framework TRIE++ for \zzcedit{extracting information from VRDs}, which can be trained from scratch, with no need for stage-wise training strategies. 
(2) We implement the framework by simultaneously learning text reading and information extraction tasks via a well-designed multi-modal context block, and also verify the mutual influence of text reading and information extraction. 
(3) To make evaluations more comprehensive and convincing, we define and divide VRDs into four categories, in which three kinds of real-life benchmarks are collected with full annotations. 
For each kind of document, we provide \zzcedit{or recommend} the corresponding benchmarks,  experimental settings, and strong baselines.  
(4) Extensive evaluations on four kinds of real-world benchmarks show superior performance compared with the state-of-the-art. Those benchmarks cover diverse types of document images, from fixed to variable layouts, from structured to semi-unstructured text types.

Declaration of major extensions compared to the conference version \cite{zhang2020trie}:  
(1) Instead of modelling context with only layout and textual features in \cite{zhang2020trie}, we here enhance the multi-modal context block by fusing three kinds of features (\ie layout, visual and textual features) with a spatial-aware attention mechanism.  
Besides, we expand the application ranges of our method, showing the ability to handle with four kinds of VRDs.  
(2) Following the suggestions in the conference reviews that the prior knowledge may be helpful to our method, we also attempt to introduce the pre-trained language model \cite{denk2019bertgrid} into the framework with a knowledge absorption module for further improving the information extraction performance.  
(3) We address the problem of performance comparison in existing methods, and then define the four categories of VRDs.  
To promote the document understanding area, we recommend the corresponding benchmarks, experimental settings, and strong baselines for each kind of document. 
(4) We explore the effects of the proposed framework with more extensive experimental evaluations \zzcedit{and comparisons}, which demonstrates its advantages. 

%% file: sections/2_related.tex
\section{Related Works}
\label{related_work} 
Thanks to the rapid expansion of artificial intelligence techniques \cite{zhuang2020next}, advanced progress has been made in many isolated applications such as document layout analysis \cite{esser2012automatic,xu2019layoutlm},  
scene text spotting \cite{liu2018fots,Qiao2020MANGOAM}, 
video understanding \cite{xu2019segregated}, 
named entities identification \cite{yadav2019survey}, 
question answering \cite{duan2018temporality}, or even causal inference \cite{kuang2020causal} etc. 
However, it is crucial to build multiple knowledge representations for understanding the complex and challenging world.
VRD \zzcedit{information extraction} is such a real task greatly helping office automation, which relies on integrating multiple techniques, including object detection, sequence learning, information extraction and even the multi-modal knowledge representation. 
Here, we roughly brief techniques as follows. 

\subsection{Text Reading} 
Text reading belongs to the OCR research field and has been widely studied for decades. 
A text reading system usually consists of two parts: text detection and text recognition.  

In \emph{text detection}, methods are usually divided into two categories: {anchor-based methods} and {segmentation-based methods}. 
Following Faster R-CNN \cite{RenHG017}, {anchor-based methods}~\cite{he2017single, liao2017textboxes,liao2018textboxes++,liao2018rotation,ma2018arbitrary,liu2017deep, shi2017detecting,Rosetta18Borisyuk} predicted the existence of texts and regress their location offsets at pre-defined grid points of the input image. 
To localize arbitrary-shaped text, Mask RCNN \cite{HeGDG17mask}-based methods \cite{xie2018scene,Zhang2019look,liu2019Towards} were developed to capture irregular text and achieve better performance.
Compared to anchor-based methods, segmentation can easily be used to describe the arbitrary-shaped text. 
Therefore, many {segmentation-based methods}~\cite{zhou2017east,long2018textsnake,Wang2019Shape,xu2019textfield} were developed to learn the pixel-level classification tasks to separate text regions apart from the background. 
In \emph{text recognition}, the encoder-decoder architecture \cite{CRNN,shi2018aster,cheng2017focusing} dominates the research field, including two mainstreaming routes: CTC\cite{Graves2006}-based \cite{shi2016end,Rosetta18Borisyuk,wang2017gated,R2AM} and attention-based \cite{cheng2017focusing,shi2018aster,cheng2018aon} methods. 
To achieve the global optimization between detection and recognition, many end-to-end trainable methods~\cite{liu2018fots,li2017towards,he2018end,busta2017deep, wang2020all,qiao2020textperceptron,feng2019textdragon,MaskTextspotter18Lyu,Qiao2020MANGOAM} were proposed, and achieved better results than the pipeline approaches. 

\zzcedit{
Global optimization is one of the importance research trend exactly 
in general object detection and text spotting. 
Hence, it is easy to observe its importance in extracting information from VRDs. Based on this, we attempt to develop an end-to-end trainable framework from text reading to information extraction.
Unlike text spotting methods achieving global optimization via a  RoI-like operations \cite{liu2018fots,feng2019textdragon} only, the proposed framework relies on an all-around module, i.e., the multi-modal context block, to bridge the vision tasks and the NLP task. This module should have the ability of both handling complex multi-modal information and providing differentiable passages among different modules.
}

\subsection{Information Extraction}
Information extraction is a traditional research topic and has been studied for many years. 
Here, we divide existing methods into two categories as follows. 

\subsubsection{Rule-based Methods}
Before the advent of learning-based models, rule-based methods\cite{riloff1993automatically,huffman1995learning,muslea1999extraction,dengel2002smartfix,schuster2013intellix,esser2012automatic} dominated this research area. 
It is intuitive that the key information can be identified by matching a predefined pattern or template in the unstructured text.
Therefore, expressive pattern matching languages \cite{riloff1993automatically,huffman1995learning} were developed to analyze syntactic sentence, and then output one or multiple target values.

To extract information from general documents such as business documents, many solutions \cite{dengel2002smartfix,schuster2013intellix,Rusiol2013FieldEF,esser2012automatic,Medvet2010APA} were developed by using the pattern matching approaches.
In detail, \cite{schuster2013intellix,Rusiol2013FieldEF,Cesarini2003AnalysisAU} required a predefined document template with relevant key fields annotated, and then automatically generated patterns matching those fields.
\cite{dengel2002smartfix,esser2012automatic,Medvet2010APA} all manually configured patterns based on keywords, parsing rules or positions.  
The rule-based methods heavily rely on the predefined template, and are limited to the documents with unseen templates. 
As a result, it usually requires deep expertise and a large time cost to conduct the templates' design and maintenance. 
 
\subsubsection{Learning-based Methods}
Learning-based methods can automatically extract key information by applying machine learning techniques to a prepared training dataset. 

Traditionally machine learning techniques like logistic regression and SVM were widely adopted in document analysis tasks. 
\cite{Shilman2005LearningNG} proposed a general machine learning approach for the hierarchical segmentation and labeling of document layout structures. 
This approach modeled document layout as grammar and performed a global search for the optimal parse based on a grammatical cost function. This method utilized machine learning to discriminatively select features and set all parameters in the parsing process.  

The early methods often ignore the layout information in the document, and then the document understanding task is downgraded to the pure NLP problem.
That is, many named entity recognition (NER) based methods ~\cite{lample2016neural,ma2019end,yadav2019survey,devlin2018bert,dai2019transformer,yang2019xlnet} can be applied to extract key information from the one-dimensional plain text. 
Inspired by this idea, ~\cite{palm2017cloudscan} proposed CloudScan, an invoice analysis system, which used recurrent neural networks to extract entities of interest from VRDs instead of templates of invoice layout. 
~\cite{sage2019recurrent} proposed a token level recurrent neural network for end-to-end table field extraction that starts with the sequence of document tokens segmented by an OCR engine and directly tags each token with one of the possible field types.  
However, they discard the layout information during the text serialization, which is crucial for document understanding.

Observing the rich {layout and} visual information contained in document images, researchers tended to incorporate more details from VRDs. Some works \cite{katti2018chargrid,denk2019bertgrid,zhao2019cutie,palm2019attend,wang2021tag} took the layout into consideration, and worked on the reconstructed character or word segmentation of the document. 
Concretely, 
\cite{katti2018chargrid} first achieved a new type of text representation by encoding each document page as a two-dimensional grid of characters. Then they developed a generic document understanding pipeline named Chargrid for structured documents by a fully convolutional encoder-decoder network.
As an extension of Chargrid, \cite{denk2019bertgrid} proposed Bertgrid in combination with a fully convolutional network on a semantic instance segmentation task for extracting fields from invoices. 
To further explore the effective information from both semantic meaning and spatial distribution of texts in documents, 
\cite{zhao2019cutie} proposed a convolutional universal text information extractor by applying convolutional neural networks on gridding texts where texts are embedded as features with semantic connotations. 
\cite{palm2019attend} proposed the attend, copy, parse architecture, an end-to-end trainable model bypassing the need for word-level labels. 
\cite{wang2021tag} proposed a tag, copy or predict network by first modelling the semantic and layout information in 2D OCR results, and then learning the information extraction in a weakly supervised manner.
{Contemporaneous with the above-mentioned methods, there are methods \cite{liu2019graph,MajumderPTWZN20,sun2021spatial,xu2019layoutlm,Xu2020LayoutLMv2MP,li2021structurallm,li2021structext} which resort to graph modeling to learn relations between multimodal inputs.}
\cite{liu2019graph} introduced a graph convolution-based model to combine textual and layout information presented in VRDs, in which graph embedding was trained to summarize the context of a text segment in the document, and further combined with text embedding for entity extraction. 
\cite{MajumderPTWZN20} presented a representation learning approach to extract structured information from templatic documents, which worked in the pipeline of candidate generation, scoring and assignment.
\cite{sun2021spatial}  modelled document images as dual-modality graphs by encoding both textual and visual features, then generated key information with the proposed Spatial Dual-Modality Graph Reasoning method (SDMG-R). Besides, they also released a new dataset named WildReceipt. 

\zzcedit{Recently, some researchers attempted to train the large pre-trained model for better results. 
\cite{xu2019layoutlm} proposed the LayoutLM to jointly model interactions between text and layout information across scanned document images. 
Then \cite{Xu2020LayoutLMv2MP} released {LayoutLMv2} 
where new model characteristics and pre-training tasks were leveraged. 
\cite{li2021structurallm} also proposed a pre-trained model and leveraged cell-level layout information instead of token-level used in \cite{xu2019layoutlm,Xu2020LayoutLMv2MP}.
\cite{li2021structext} proposed a unified framework to handle the entity labeling and entity linking sub-tasks, using token and segment level context.
\cite{garncarek2021lambert} proposed LAMBERT by modifying the Transformer encoder architecture for obtaining layout features  from an OCR system, without the need to re-learn language semantics from scratch.
\cite{powalski2021going} proposed TILT which relies on  a pretrained encoder-decoder Transformer to learns layout information, visual features, and textual semantics. 
\cite{li2021selfdoc} developed SelfDoc by exploiting the positional, textual, and visual information of a document and modeling their contextualization.
However, these works rely on the big models trained on the large datasets (\eg IIT-CDIP \cite{Lewis2006BuildingAT}), which suffers from the computational cost problem in real applications.
On the other hand, they also ignore the mutual impacts between text reading and information extraction modules.
They need to recompute visual features used in IE module, leading to double-computing problems, and cannot exploit supervision from IE on text reading.
}

\subsection{End-to-End Information Extraction from VRDs}
Two related concurrent works were presented in~\cite{qin2019eaten,carbonell2019treynet}.
\cite{qin2019eaten} proposed an entity-aware attention text extraction network to extract entities from VRDs.
However, it could only process documents of relatively fixed layout and structured text, like train tickets, passports and business cards.
\cite{carbonell2019treynet} localized, recognized and classified each word in the document.
Since it worked in the word granularity, it required much more labeling efforts (layouts, content and category of each word) and  had difficulties  extracting those entities which were embedded in word texts (\eg extracting `51xxxx@xxx.com' from  `153-xxx97$|$51xxxx@xxx.com').
Besides, in its entity recognition branch, it still worked on the serialized word features, which were sorted and packed in the left to right and top to bottom order. 
The two existing works are strictly limited to documents of relatively fixed layout and one type of text (structured or semi-structured). 
Similar to the conference version \cite{zhang2020trie} of our method, \cite{wang2021towards} recently proposed an end-to-end \zzcedit{information extraction} framework accompanied by a Chinese examination paper head dataset. 
Unlike them, our method acts as a general \zzcedit{VRD information extraction framework}, and can handle documents of both fixed and variable layouts, structured and semi-structured text types.

%% file: sections/3_approach.tex
\section{Methodology}\label{methodology}
\begin{figure}
	\centering
	\includegraphics[width=0.49\textwidth]{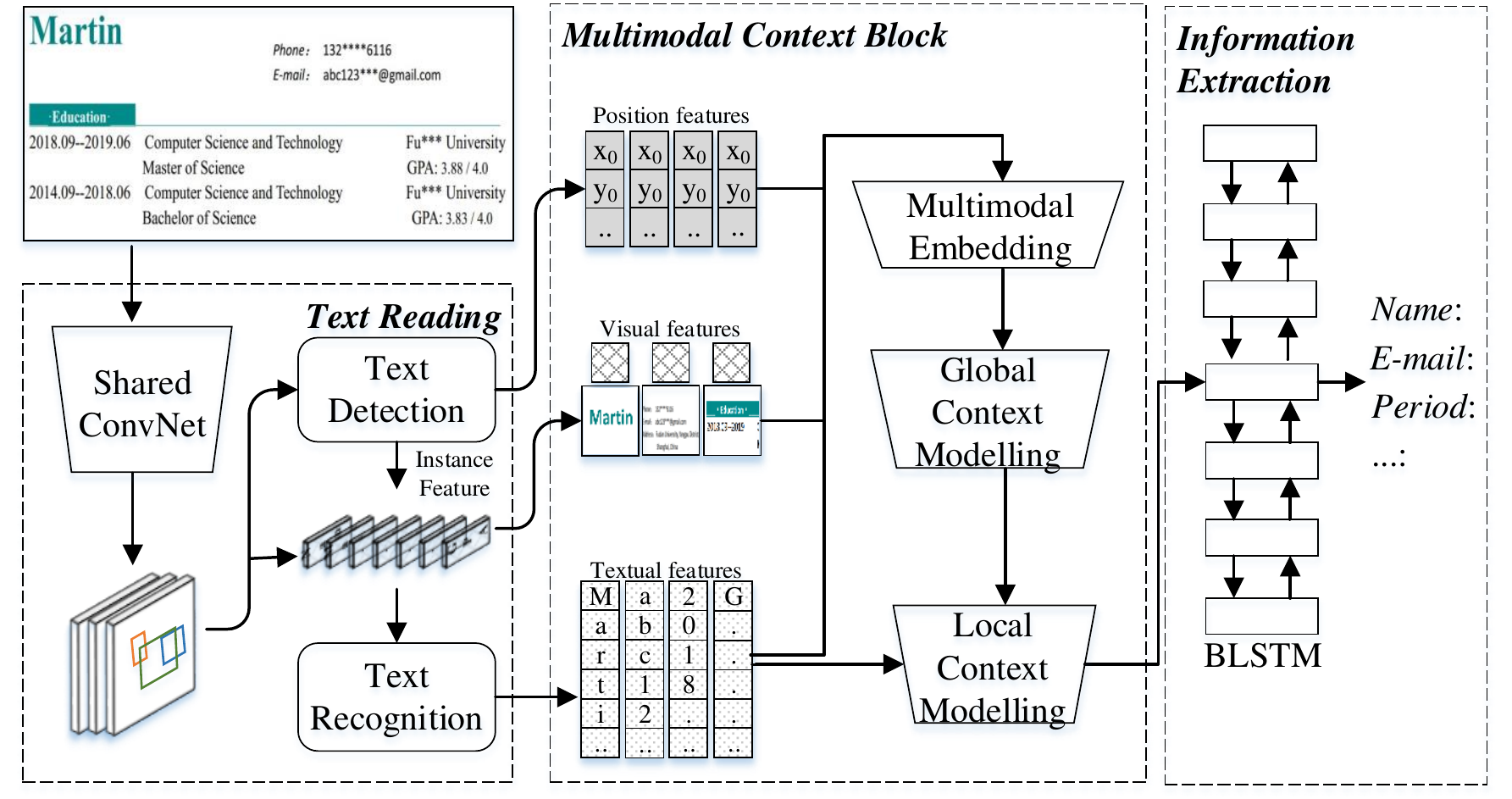}
	\caption{The overall framework. The network predicts text locations, text contents and key entities in a single forward pass.
}
	\label{fig:system_architecture}
\end{figure}
This section introduces the proposed framework, which has three parts: text reading, multi-modal context  block and information extraction module, as shown in Figure~\ref{fig:system_architecture}. 


\subsection{Text Reading}
Text reading module commonly includes a shared convolutional backbone, a text detection branch as well as a text recognition branch.
We use ResNet-D~\cite{he2019bag} and Feature Pyramid Network (FPN) \cite{LinDGHHB17feature} as our backbone to extract the shared convolutional features. 
For an input image $x$, we denote $\mathcal{I}$ as the shared feature maps. 

\textbf{Text detection}. 
The branch takes $\mathcal{I}$ as input and predicts the locations of all candidate text regions, i.e.,
\begin{equation}\label{equa1}
\mathcal{B}=\textit{Detector}(\mathcal{I})
\end{equation}
where the $\textit{Detector}$ {can be the anchor-based~\cite{he2017single,liao2017textboxes,liu2017deep,shi2017detecting} or segmentation-based ~\cite{zhou2017east,long2018textsnake,Wang2019Shape} text detection heads.}
$\mathcal{B}=(b_1, b_2,\dots, b_m)$ is a set of $m$ text bounding boxes, and $b_i=(x_{i0}, y_{i0}, $ $x_{i1}, y_{i1})$ denotes the top-left and bottom-right positions of the $i$-th text. 
In mainstream methods, RoI-like operations (\emph{e.g.}, RoI-Pooling \cite{RenHG017} used in \cite{li2017towards}, ROI-Align \cite{HeGDG17mask} used in \cite{he2018end}, RoI-Rotate used in \cite{liu2018fots}, or even RoI-based arbitrary-shaped transformation \cite{qiao2020textperceptron,wang2020all}) are applied on the shared convolutional features $\mathcal{I}$ to get their text instance features. 
Here, the text instance features are denoted as $\mathcal{C}=(c_1, c_2,\dots, c_m)$. 
The detailed network architecture is shown in Section \ref{sec-impl}. 

\textbf{Text recognition}. 
The branch predicts a character sequence from each text region features $c_i$.
Firstly, each instance feature $c_i$ is fed into an encoder (\eg CNN and LSTM \cite{LSTM}) to extract a higher-level feature sequence $\mathcal{H}=(h_1, h_2, \dots, h_l)$, where $l$ is the length of the extracted feature sequence.
Then, a general sequence decoder (\eg attention-based \cite{shi2016end,cheng2017focusing}) is adopted to generate the sequence of characters $y=(y_1, y_2,\dots, y_T)$, where $T$ is the length of label sequence. 
Details are shown in Section \ref{sec-impl}. 

We choose {attention-based sequence decoder} as the character recognizer. It is a recurrent neural network that directly generates the character sequence $y$ from an input feature sequence $\mathcal{H}$.
\zzcedit{
To generate the $t$-th character, the attention weight $\alpha_t$ and the corresponding glimpse vector $g_t$ are computed as 
$g_t=\sum_{k=1}^{l}\alpha_{t,k}{h}_k$
where 
$\alpha_{t,k}=exp(e_{t,k})/\sum_{j=1}^{l}exp(e_{t,j})$. 
Here  
$e_{t,j}=w^\mathsf{T}tanh(W_ss_{t-1}+W_h{h}_j+b)$, 
and $w$, $W_s$, $W_h$ and $b$ are trainable weights. 
Then the hidden state $s_{t-1}$ is updated via,
\begin{equation}
s_t=LSTM(s_{t-1}, g_t, y_{t-1}).
\end{equation} 
Finally, $s_t$ is taken for predicting the current-step character, i.e.,
$p(y_t)=softmax(W_{o}s_t + b_{o})$
where both $W_{o}$ and $b_{o}$ are learnable weights.
}

\subsection{Multi-modal Context Block}
We design a multi-modal context block to consider layout features, visual features and textual features altogether. 
Different modalities of information are complementary to each other, and fully fused for providing robust multi-modal feature representation. 

\subsubsection{Multi-modal Feature Generation} 
Document details such as the apparent color, font, layout and other informative features also play an important role in document understanding. 

A natural way of capturing the layout and visual features of a text is to resort to the convolutional neural network. 
Concretely, 
the position information of each text instance is obtained from the detection branch, i.e., $\mathcal{B}=(b_1, b_2,\dots, b_m)$. 
For visual feature, different from \cite{xu2019layoutlm,Xu2020LayoutLMv2MP} which extract these features from scratch, we directly reuse text instance features $\mathcal{C}=(c_1, c_2, \dots, c_m)$ by text reading module as the visual features. 
Thanks to the deep backbone and lateral connections introduced by FPN, each  $c_i$ summarizes the rich local visual patterns of the $i$-th text.

In sequence decoder, give the $i$-th text instance, its represented feature of characters before softmax contain rich semantic information.  
For the attention-based decoder, we can directly use $z_i=(s_1, s_2, \dots, s_T)$ as its textual features. 

\subsubsection{Prior Knowledge Absorption} 
\begin{figure}
	\centering
	\includegraphics[width=0.45\textwidth]{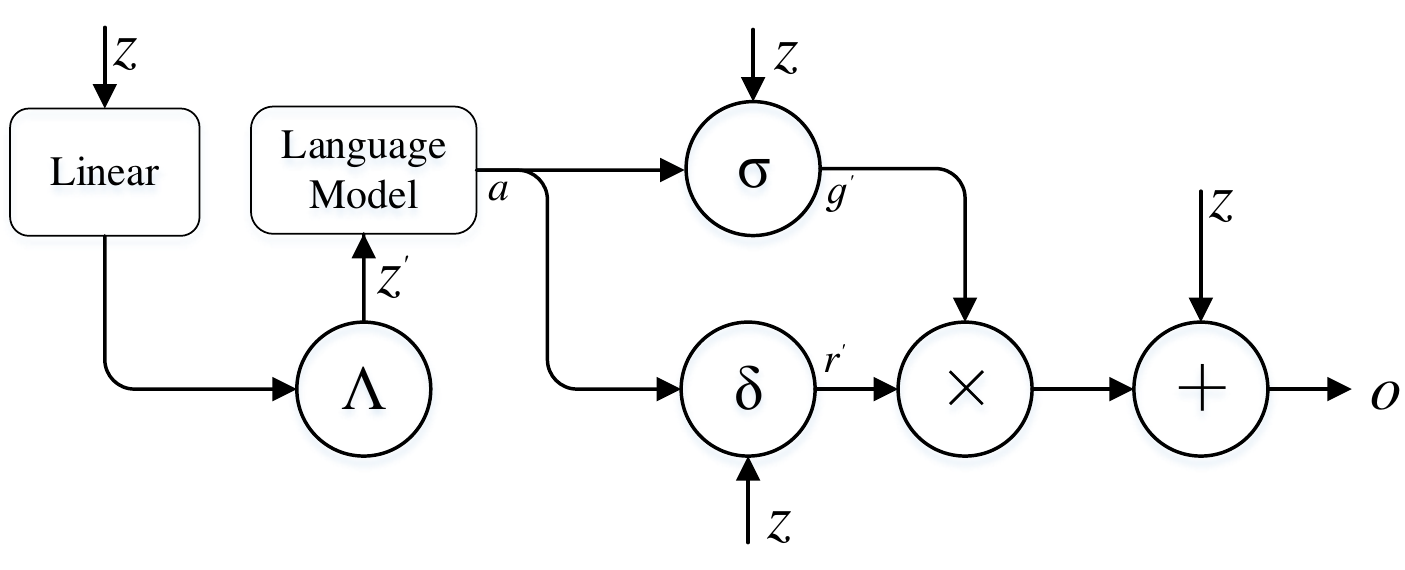}
	\caption{\zzcedit{The knowledge absorption mechanism. $\sigma$ and $\delta$ separately mean the Sigmoid and Tanh operations. $\times$ and $+$ refer to the element-wise multiplication and addition, respectively. $\wedge$ means the argmax operation.
	}
}
	\label{fig:prior}
\end{figure}

Since pre-trained language model contains general language knowledge like semantic properties, absorbing knowledge from the language model may help improve the performance of information extraction.
Compared to the conference paper \cite{zhang2020trie}, we here attempt to bring the language model into our framework. 
{However, prior language information has different contributions on different VRDs. 
For example, on Resume scenario that require semantics, prior language information contributes more, while on Taxi scenario which requires less semantics, prior language information contributes less.}
%
Inspired by the gating operation in LSTM \cite{LSTM}, we design a gated knowledge absorption mechanism to adjust the prior knowledge flows in our framework, as shown in Figure \ref{fig:prior}. 

\zzcedit{
Concretely, we first transform textual input $z$ into vocabulary space via a Linear layer, then obtain the character sequence by conducting argmax operation ($\wedge$), i.e.,
\begin{equation}
\label{z-prime}
z^\prime = \wedge(Linear(z)). 
\end{equation}
Then $z^\prime$ is fed into the language model, outputting its knowledge representation $a$.
} 

In order to dynamically determine the degree of dependency of the pre-trained  model, we use an on-off gate $g^\prime $ 
\begin{equation}
g^\prime = \sigma(W_{g^\prime}a + U_{g^\prime}z + b_{g^\prime})
\end{equation}
to balance the flow of the prior knowledge activation $r^\prime$ 
\begin{equation}
r^\prime = \delta(W_{r^\prime}a + U_{r^\prime}z + b_{r^\prime}).
\end{equation}
Here, the gate is used for determining whether general knowledge is needed.  
Then the modulated textual feature $o$ is calculated as 
\begin{equation}
\label{gating}
o = g^\prime \odot r^\prime + W_oz.
\end{equation}

\subsubsection{Multi-modal Context Modelling}
We first embed each modality information into feature sequences with the same dimension, and fuse them with a normalization layer. 
Inspired by the powerful Transformer \cite{devlin2018bert,VisualBERTLi,Lu2019ViLBERT,Xu2020LayoutLMv2MP}, the self-attention mechanism is used to build deep relations among different modalities, \zzcedit{whose input  consists of queries $Q$ and keys $K$ of dimension $d_k$, and values $V$ of dimension $d_v$.} 


\textbf{Multi-modal Feature Embedding} 
Given a document with $m$ text instance, we can capture the inputs of position $\mathcal{B}=(b_1,b_2,\dots,b_m)$, the inputs of visual feature $\mathcal{C}=(c_1,c_2,\dots,c_m)$ and the inputs of modulated textual feature $o=(o_1,o_2,\dots,o_m)$. 

Since position information provides layout information of documents, we introduce a position embedding layer to preserve layout information, for the $i$-th text instance in a document,
\begin{equation}
pe_i=\sum_{j=1}^{|b_i|} embedding(b_{ij}),
\end{equation}
where $embedding$ is a learnable embedding layer, $b_i=(x_{i0},y_{i0},x_{i1},y_{i1})$ and $pe_i\in \mathbb{R}^{d_e}$.

For $c_i$ visual feature, we embed it using a convolutional neural network layer with the same shape of $pe_i$,
\begin{equation}
\widehat{c_i}=ConvNet_c(c_i).
\end{equation}

For $o_i$ textual feature, a $ConvNet$ of multiple kernels similar to~\cite{zhang2015character} is used to aggregate semantic character features in $o_i$ and outputs $\widehat{z_i}\in\mathbb{R}^{d_e}$,
\begin{equation}
\widehat{z_i}=ConvNet_z(o_i). 
\label{eq:textual}
\end{equation}

Then, the $i$-th text's embedding is fused of $\widehat{c_i}$, $\widehat{z_i}$ and $pe_{i}$, followed by the {$LayerNorm$} normalization, defined as
\begin{equation}
emb_i=LayerNorm(\widehat{c_i} + \widehat{z_i} + pe_i). 
\end{equation}
Afterwards, we pack all the texts' embedding vector together, i.e., $emb=(emb_1, emb_2, \dots, emb_m)$, which serves as the $K$, $Q$ and $V$ in the scaled dot-product attention.


\textbf{Spatial-Aware Self-Attention} 
{
To better learn pair-wise interactions between text instances, we use the spatial-aware self-attention mechanism instead of the original self-attention, and the correlative context features
}
$\widetilde{\mathcal{C}}=(\widetilde{c_1}, \widetilde{c_2}, \dots, \widetilde{c_m})$  are obtained by,
\begin{equation}
\begin{split}
\widetilde{\mathcal{C}}&=Attention(Q,K,V) \\
&=softmax(\frac{QK^\mathsf{T}}{\sqrt{d_{info}}}+pe_{\Delta \mathcal{B}})V
\end{split}
\end{equation}
where $d_{info}$ is the dimension of text embedding, and $\sqrt{d_{info}}$ is the scaling factor. 
$pe_{\Delta \mathcal{B}}$ refers to the spatial-aware information, and is calculated by embedding features of position relations ${\Delta \mathcal{B}}$ among different text instances in $\mathcal{B}$, i.e., $pe_{\Delta \mathcal{B}}= embedding({\Delta \mathcal{B}})$. Here, ${\Delta \mathcal{B}}$ is defined as 
\begin{equation}
\Delta \mathcal{B} = 
\left[
\begin{array}{cccc}
0 			& b_1-b_2 	& \cdots 	& b_1-b_m\\
b_2-b_1 	& 0 			& 	\cdots	& b_2-b_m\\
\cdots 	& \cdots 	& \cdots		&\cdots \\
b_m-b_1 	& b_m-b_2	& 	\cdots	& 0
\end{array}
\right].
\end{equation}
To further improve the representation capacity of the attended feature, multi-head attention is introduced. Each head corresponds to an independent scaled dot-product attention function and the text context features $\widetilde{\mathcal{C}}$ is given by:
\begin{equation}
\begin{split}
\widetilde{\mathcal{C}}&=MultiHead(Q,K,V)\\
&=[head_1, head_2, ..., head_n]W^{info}
\end{split}
\end{equation}
\begin{equation}
head_j=Attention(QW_j^Q, KW_j^K, VW_j^V)
\end{equation}
where $W^Q_j$, $W^K_j$ and $W^V_j$ $\in \mathbb{R}^{(d_{info}\times d_n)}$ are the learned projection matrix for the $j$-th head, $n$ is the number of heads, and $W^{info}\in \mathbb{R}^{(d_{info} \times d_{info})}$. To prevent the multi-head attention model from becoming too large, we usually have $d_n = \frac{d_{info}}{n}$.

\textbf{Context Fusion}
{
Both the multi-modal context and textual features matter in entity extraction. The multi-modal context features ($\widetilde{\mathcal{C}}$) provide necessary information to tell entities apart while the textual features $o$ enable entity extraction in the character granularity, as they contain semantic features for each character in the text.
Thus, we need to fuse them further.
}
That is, for the $i$-the text instance, we pack its multi-modal context vector $\widetilde{c_i}$ and its modulated textual features $o_i$ together along the channel dimension, i.e., 
$(u_{i1}, u_{i2},\dots, u_{iT})$ where $u_{ij}=[o_{i,j}, c_i]$. 

\subsection{Information Extraction} \label{ie}
Then, a Bidirectional-LSTM is applied to further model the long dependencies within the characters,
\begin{equation}
H_{i}^\prime=(h_{i,1}^\prime, h_{i,2}^\prime, \dots, h_{i,T}^\prime) = BiLSTM(u_i),
\end{equation}
which is followed by a fully connected network and a \zzcedit{conditional random field (\emph{abbr}, CRF)}  layer, projecting the output to the dimension of \zzcedit{IOB\footnote{
\zzcedit{Here, `I' stands for `inside', signifying that the word is inside a named entity.
`O' stands for `outside', referring to the word is just a regular word outside of a named entity.
`B' stands for `beginning', signifying beginning of a named entity. 
}
}} \cite{SangV99representing} label space.
\begin{equation}
p_{i,j}^{info} = CRF(Linear(h_{i,j}^\prime))
\end{equation}

\zzcedit{Note that, unlike the spatial-aware self-attention learning pair-wise interactions  between text instances (\emph{e.g,} the `Phone Number' and `Name'), BiLSTM is for depicting the relations among the characters in each instance (\emph{e.g,} the numbers in `Phone Number' or characters in `Name').}

\subsection{Optimization}\label{sec3.5}
The proposed network can be trained in an end-to-end manner and the losses are generated from three parts,
\begin{equation}
\label{losses}
\mathcal{L}=\mathcal{L}_{det} + \lambda_{recog}\mathcal{L}_{recog} + \lambda_{info}\mathcal{L}_{info}
\end{equation}
where hyper-parameters $\lambda_{recog}$ and $\lambda_{info}$ control the trade-off between losses.

$\mathcal{L}_{det}$ is the loss of text detection branch, which can be formulated as different forms according to the selected detection heads. Taking Faster-RCNN \cite{RenHG017} as the detection head, the detection part consists of a classification loss and a regression loss. 

For sequence recognition part, 
the attention-based recognition loss is 
\begin{equation}
\mathcal{L}_{recog}=-\frac{1}{T}\sum_{i=1}^{m}\sum_{t=1}^{T}log\ p(\hat{y}_{i,t}|\mathcal{H}),
\end{equation}
where $\hat{y}_{i,t}$ is the ground-truth label of $t$-th character in $i$-th text from recognition branch. 

The information extraction loss {is the CRFLoss, as used in~\cite{lample2016neural,wang2021towards}.}
\zzcedit{
Specifically, for the $i$-th text instance, the input sequence of LSTM is $u_i=(u_{i1}, u_{i2},\dots, u_{iT}) \in \mathbb{R}^{T \times d_{in}}$, while the output is $H^\prime_i \in \mathbb{R}^{T \times \Omega}$, where $\Omega$ denotes the number of entities.
$H^\prime$ (we omit the subscript $i$ for convenience) denotes the scores of emissions matrix, and $H^\prime_{t,j}$ represents the score of the $j-$th entity of the $t-$th character in $u_i$.
For a sequence of predictions $y=(y_1, y_2,\dots, y_T)$, its scores can be defined as follows
\begin{equation}
s(u_i, y)=\sum_{j=0}^{T}A_{y_j,y_{j+1}}+\sum_{j=1}^{T}H^\prime_{j,y_j},
\end{equation}
where $A$ is a matrix of transition scores such that $A_{i,j}$ represents the score of a transition from the tag $i$ to tag $j$. 
$y_0$ and $y_T$ are the start and end tags of a sentence, that we add to the set of possible tags. $A$ is therefore a square matrix of size $\Omega+2$.
A softmax over all possible tag sequences yields a probability for the sequence $y$, i.e., 
$p(y|u_i)=\frac{e^{s(u_i,y)}}{\sum_{\widetilde{y} \in \mathbb{Y}(u_i)}e^{s(u_i,\widetilde{y})}}$
where $\mathbb{Y}(u_i)$ is all possible entity sequences for $u_i$.

During training, we minimize the negative log-likelihood estimation of the correct entity sequence as:
\begin{equation}
\mathcal{L}_{info}=-log(p(y|u_i))=-s(u_i,y)+log(\sum_{\widetilde{y} \in \mathbb{Y}(u_i)}e^{s(u_i, \widetilde{y})}).
\end{equation}
While decoding, we predict the output sequence that obtains the maximum score given by 
$y^*=\mathop{\arg\max}\limits_{\widetilde{y} \in \mathbb{Y}(u_i)} p(\widetilde{y}|u_i)$.
}

Note that since \textit{text reading} and \textit{information extraction} modules are bridged with the multi-modal context block, they can reinforce each other.
Specifically, the multi-modality features of text reading are fully fused and essential for information extraction. 
At the same time, the semantic feedback of information extraction also contributes to the optimization of the shared convolutions and text reading module.

%% file: sections/4_benchmarks.tex
\begin{table*}
	\begin{center}
		\caption{Statistics of popular VRD benchmarks. 
\emph{pos, text} and \emph{entity} mean the word position, character string and entity-level annotations, respectively. 
Term `Syn' and `Real' separately mean the dataset is generated by a synthetic data engine and manual collection. 
The new collected datasets are available at {\emph{https://davar-lab.github.io/dataset/vrd.html}}.
}
		\label{table:datasets}
		\begin{tabular}{clccccccc}
			\toprule
			Category & Dataset    &  \#Training  & \#Validation  &  \#Testing  & \#Entity & \#Instance  & Annotation Type & Source\\ 
			\midrule
			\makecell[c]{I} 
				&Train ticket \cite{qin2019eaten} & 271.44k & 30.16k&400 &  5 & -  & [\emph{entity}] & Syn\\ 
				&Passport \cite{qin2019eaten} & 88.2k &9.8k&2k &  5 & -  &  [\emph{entity}] & Syn\\
				&Taxi Invoice & 4000 & -&1000  &  9 & 136,240 &  [\emph{pos, text, entity}]& Real \\	
			\midrule
			\makecell[c]{II}
				&Business Email & 1146 & - &499 & 16 & 35,346 &  [\emph{pos, text, entity}]& Real\\
			\midrule
			\makecell[c]{III}
				&SROIE \cite{HuangCHBKLJ19competition} & 626 & - &347 &  4 &  52,451 & [\emph{pos, text, entity}]&Real\\	
				&Business card \cite{qin2019eaten} & 178.2k &19.8k &2k &  9 & -  &  [\emph{entity}]&Syn\\
				&FUNSD \cite{Jaume2019FUNSDAD} & 149 & -&50 &  4 & 31,485 &   [\emph{pos, text, entity}]&Real\\
				&CORD$^3$ \cite{Park2019CORDAC} & 800 & 100&100 &  30 & - &  [\emph{pos, text, entity}]&Real\\
				&EPHOIE \cite{wang2021towards} & 1183 & - &311 &  12 & 15,771 &   [\emph{pos, text, entity}]& Real\\
				&WildReceipt \cite{sun2021spatial} &1267  & - & 472& 26  & 69,000 &   [\emph{pos, text, entity}]& Real\\
			\midrule
			\makecell[c]{IV}
				&Kleister-NDA\cite{Gralinski2020KleisterAN} & 254 & 83&203 &  4 & - &[\emph{pos, text, entity}]& Real\\
				&Resume & 1045 & -&482 &  11 & 82,800 &   [\emph{pos, text, entity}]& Real \\
			\bottomrule
		\end{tabular}
	\end{center}
\end{table*}

\section{Benchmarks} \label{benchmark}

As addressed in Section 1, most existing works verify their methods on private datasets due to their privacy policies. 
It leads to difficulties for fair comparisons between different approaches. 
Though existing datasets like SROIE \cite{HuangCHBKLJ19competition} have been released, they mainly fall into Category III, i.e., documents with variable layout and structured text type.
The remaining three kinds of application scenarios (Category I, II and IV) have not been studied well because of the limited real-life datasets.

\subsection{Dataset inventory} 
To boost the research of VRD understanding, we here extend the benchmarks of VRD, especially on Category I, II and IV. 
Table \ref{table:datasets} shows the detailed statistics of these benchmarks.
\begin{itemize}
\item \emph{Category I} refers to document images with uniform layout and structured text type, which is very common in everyday life. Contrastively, its research datasets are very limited due to various privacy policies. 
Here, we find only two available benchmarks, i.e., train ticket and passport dataset released by \cite{qin2019eaten}, which are generated with a synthetic data engine and provide only entity-level annotations. 
To remedy this issue, we release a new real-world dataset containing 5000 taxi invoice images. Except for providing the text position and character string information for OCR tasks (text detection and recognition),  entity-level labels including 9 entities ({Invoice code, Invoice number, Date, Get-on time, Get-off time, Price, Distance, Wait time, Total}) are also provided.
{Besides, this dataset is very challenging, as many images are in low-quality (such as blur and occlusion).} 
\item \emph{Category II} refers to those documents with fixed layout and semi-structured text type, like business email or national housing contract. 
NER datasets like CLUENER2020 \cite{xu2020cluener2020} are only collected for NLP tasks, and they provide only semantic content while ignoring the important layout information. As addressed in Section \ref{sec:introduction}, the joint study of OCR and IE is essential. 
Unfortunately, we have not found available datasets that contains both OCR and IE annotations.  
We also ascribe the issue to various privacy policies.
We here collect a new business email dataset from RVL-CDIP \cite{Harley2015EvaluationOD}, which has 1645 email images with  35346 text instances and 15 entities ({To, From, CC, Subject, BCC, Text, Attachment, Date, To-key, From-key, CC-key, Subject-key, BCC-key, Attachment-key, Date-key}).
\item \emph{Category III} means documents which are with variable layout and structured text type like purchase receipt dataset SROIE \cite{HuangCHBKLJ19competition}. These datasets are usually composed of small documents (\emph{e.g.}, purchase receipts, business cards, etc.), and entities are organized in a predetermined schema. 
We note that most previous literature focus on this category.  
We here list five available datasets. 
SROIE is a scanned receipt dataset widely evaluated in many methods, which is fully annotated and provides text position, character string and key-value labels.  
Business card is a synthesized dataset released by \cite{qin2019eaten}, and has only key-value pair annotations without OCR annotations. 
FUNSD \cite{Jaume2019FUNSDAD} is a dataset aiming at extracting and structuring the textual content from noisy scanned forms. It has only 199 forms with four kinds of entities, i.e., question, answer, header and other.   
CORD$^2$ \cite{Park2019CORDAC} is a consolidated receipt dataset, in which images are with text position, character string and multi-level semantic labels.  
EPHOIE \cite{wang2021towards} is a Chinese examination paper head dataset, in which each image is cropped from the full examination paper. This dataset contains handwritten information, and is also fully annotated. 
WildReceipt \cite{sun2021spatial} is a large receipt dataset collected from document images of unseen templates in the wild. It contains 25 key information categories, a total of about 69000 text boxes.
\item \emph{Category IV} means documents that have variable layout and semi-structured text type.
Different from those datasets in Category III, Kleister-NDA\cite{Gralinski2020KleisterAN} aims to understand long documents (i.e., Non-disclosure Agreements document), but it provides only 540 documents with four general entity classes. 
To enrich benchmarks in this category, we release a large-scale resume dataset, which has 1527 images with ten kinds of entities({Name, Time, School, Degree, Specialty, Phone number, E-mail, Birth, Title, Security code}). 
Since resumes are personally designed and customized, it is a classic document dataset with variable layouts and semi-structured text. 
\end{itemize}

\footnotetext[3]{In the original paper, CORD claim that 11,000  receipt images were collected with 54 sub-classes. Up to now, only 1000 images have been released in total.}

\subsection{Challenges in different kinds of documents}
It will be the most straightforward task to extract entities from documents in Category I, which attributes to its complete fixed layout and structured text type. 
For this kind of documents, challenges are mainly from the text reading part, such as the distorted interference. The standard object detection methods like Faster-RCNN \cite{RenHG017} also can be further developed to handle this task. 
In Category II, the layout is fixed, but the text is semi-structured. 
Thus, in addition to modelling layout information, we also should pay  attention to mining textual information. 
Then some NLP techniques like the pre-trained language model can be exploited. 
As to the text reading part, long text recognition is also challenging.
Documents in Category III face the problem of complex layout. 
Thus the layout modelling methods \cite{liu2019graph,PICK2020YU} like graph neural networks are widely developed for coping with this issue. 
The documents in Category IV are in the face of both complex layout and NLP problems, which becomes the most challenging task.

\zzcedit{
We hope that the combed benchmarks can promote the increasingly important research topic. 
We want validate our model as well as some state-of-the-arts on the representative datasets, and also provide a group of solid baselines (See Section \ref{baseline}) for this research community. 
}

%% file: sections/5_experiments.tex
\section{Experiments}
\label{experiment}
In subsection \ref{sec-impl}, we first introduce the implementation details of network and training skills. 
In subsection \ref{ablation}, we perform ablation study to verify the effectiveness of the proposed method on four kinds of VRD datasets, i.e., Taxi Invoice, Business Email, WildReceipt and Resume. 
In subsection \ref{sota}, we compare our method with existing approaches on  several recent datasets like FUNSD, SROIE, EPHOIE and WildReceipt, demonstrating the advantages of the proposed method. 
{Then, }
we provide a group of strong baselines on four kinds of VRDs in subsection \ref{baseline}. 
Finally, we discuss the challenges of the different categories of documents. 
Codes and models are available at \emph{https://davar-lab.github.io/publication/trie++.html}. 


\subsection{Implementation Details} \label{sec-impl}
\subsubsection{Data Selecting}
To facilitate end-to-end document understanding (\textit{text reading} and \textit{information extraction}), datasets should have position, text and entity annotations.
Hence, we only consider those datasets which satisfy the above requirement.
On the ablation and strong baseline experiments, we select one classic dataset from each category, which has the largest number of samples.
They are Taxi Invoice dataset from Category I, Business Email dataset from Category II, WildReceipt dataset from Category III and Resume dataset from Category IV.
When compared with the state-of-the-arts, since they mainly report their results on popular SROIE, FUNSD and EPHOIE benchmarks, we also include these benchmarks in Section~\ref{sota}.

\subsubsection{Network Details}
The backbone of our model is ResNet-D~\cite{he2019bag}, followed by the FPN \cite{LinDGHHB17feature} to further enhance features.
The text detection branch in \textit{text reading module} adopts the Faster R-CNN \cite{RenHG017} network and outputs the predicted bounding boxes of possible texts for later sequential recognition.
For each text region, its features are extracted from the shared convolutional features by RoIAlign \cite{HeGDG17mask}. 
The shapes are represented as $32\times256$ for Taxi Invoice and WildReceipt, and $32\times512$ for Business Email and Resume. 
Then, features are further decoded by LSTM-based attention \cite{cheng2017focusing}, where the number of hidden units is set to 256. 


{In the \textit{multimodal context block}, BERT~\cite{devlin2018bert} is used as the pre-trained language model.
Then, convolutions of four kernel size $[3, 5, 7, 9]$ followed by max pooling are used to extract final textual features. 
\zzcedit{The dimension of Linear layer in Equation \ref{z-prime} is set as $\Omega \times 256$, where $\Omega$ denotes the vocabulary space.}
}

In the \textit{information extraction module}, the number of hidden units of BiLSTM used in entity extraction is set to 128.
Hyper-parameters $\lambda_{recog}$ and $\lambda_{info}$ in Equation \ref{losses} are all empirically set to 1 in our experiments.

\subsubsection{Training Details}
{Our model and its counterparts are implemented under the PyTorch framework \cite{paszke2019pytorch}.
For our model, the AdamW \cite{loshchilov2017decoupled} optimization is used. 
We set the learning rate to 1e-4 at the beginning and decreased it to a tenth at 50, 70 and 80 epochs.
The batch size is set to 2 per GPU.
For the counterparts, we separately train text reading and information extraction tasks until they are fully converged.
All the experiments are carried out on a workstation with 8 NVIDIA A100 GPUs.
}

\subsubsection{Evaluation Protocols}\label{protocals}
We also note that different evaluation protocols are adopted in previous works. 
For example in the evaluation of information extraction part, 
both EATEN \cite{qin2019eaten} and PICK \cite{PICK2020YU} used the defined mean entity accuracy (mEA) and mean entity f1-score (mEF) as metrics.  
CUTIE \cite{zhao2019cutie} adopted the average precision (AP) as the metric, and 
Chargrid \cite{katti2018chargrid} developed new evaluation metric like word error rate for evaluation. 
While the majority of methods \cite{zhang2020trie,Gralinski2020KleisterAN,xu2019layoutlm} used the F1-score as the evaluation metric. 
As a result, the non-uniform evaluation protocols bring extra difficulties on  comparisons. 
Therefore, we attempt to describe a group of uniform evaluation protocols for VRD understanding by carefully analyzing previous methods, including the evaluation protocols of text reading and information extraction parts.  

Text reading falls into the OCR community, and it has uniform evaluation standards by referring to mainstream text detection \cite{liao2017textboxes,liu2019Towards,liu2018fots} and text recognition \cite{CRNN,shi2018aster,cheng2017focusing} methods. 
\emph{precision} (\emph{abbr}. PRE$_d$) and \emph{recall} (\emph{abbr}. REC$_d$) are used to measure  performance of text localization,  and \emph{F-measure} (\emph{abbr}. F$_d$-m) is the harmonic average of \emph{precision} and \emph{recall}. 
To evaluate text recognition, the \emph{accuracy} (abbr. ACC) used in \cite{CRNN,shi2018aster,cheng2017focusing} is treat as its measurement metric.
When evaluating the performance of end-to-end text detection and recognition, the end-to-end level evaluating metrics like precision (denoted by PRE$_r$), recall (denoted by REC$_r$) and  F-measure (denoted by F$_r$-m)  following \cite{2011End} without lexicon is used, in which  all detection results are considered with an IoU$>$0.5.

For information extraction, we survey the evaluation metrics from recent research works \cite{zhang2020trie,Gralinski2020KleisterAN,xu2019layoutlm,Jaume2019FUNSDAD,liu2019graph,wang2021towards,Xu2020LayoutLMv2MP}, and find  that the precision, recall and F1-score of entity extraction are widely used. 
Hereby, we recommend the \emph{entity precision} (abbr. ePRE), \emph{entity recall} (abbr. eREC) and \emph{entity F1-score} (eF1) as the evaluation metrics for this task.

\subsection{Ablation Study} \label{ablation}
In this section, we perform the ablation study on Taxi Invoice, Business Email, WildReceipt and Resume datasets to verify the effects of different components in the proposed framework. 

\subsubsection{Effects of multi-modality features}\label{forward_effect}
\begin{figure*}[h]
	\centering
	\includegraphics[width=\textwidth]{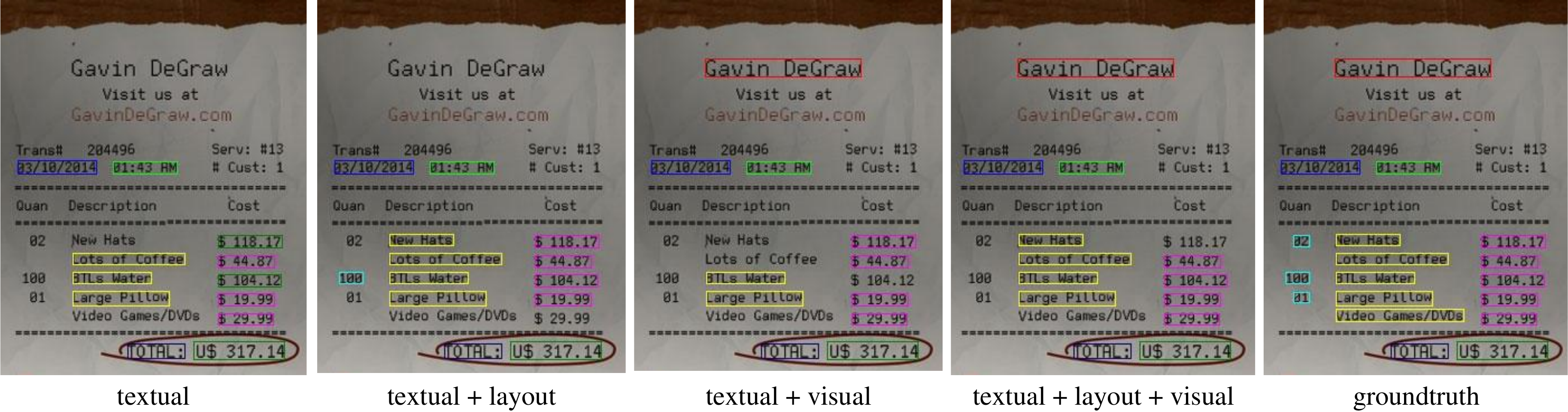}
	\caption{{Illustration of modality contributions.  Different colors denote different entities, such as {\color{red}{Store-Name}}, {\color{blue}{Date}}, {\color{green}{Time}}, {\color{yellow}{Product-Item}}, {\color{cyan}{Product-Quantity}}, {\color{VioletRed}{Product-Price}}, {\color{ForestGreen}{Total}}. Best viewed in color.}}
	\label{fig:modality_contribution}
\end{figure*}

To examine the contributions of visual, layout and textual features to information extraction, we perform the following ablation study on four kinds of datasets, and the results are shown in Table~\ref{table:performance-contribution}. 
\emph{Textual feature} means that entities are extracted using features from the text reading module only.
Since the layout information is completely lost, this method presents the worst performance.
Introducing either the \emph{visual features} or \emph{layout features} brings significant performance gains.
Further fusion of the above multi-modality features gives the best performance, which verifies the effects.
We also show examples in Figure.~\ref{fig:modality_contribution} to verify their effects. 
By using the \textit{textual feature} only, the model misses the `Store-Name' and has confusion between `Total' and `Product-Price' entities. 
Combined with the \textit{layout feature}, the model can recognize `Product-Price' correctly. 
When combined with the \textit{visual feature}, the model can recognize Store-Name, because the \textit{visual feature} contains obvious visual clues such as the large font size. 
It shows the best result by integrating all modality features.
\begin{table}[h]
	\begin{center}
		\caption{Accuracy results (eF1) with multi-modal features on information extraction.}
		\label{table:performance-contribution}
		\begin{tabular}{lcccc}
			\toprule
			Textual feature & $\surd$ & $\surd$& $\surd$& $\surd$\\
			Layout feature & & $\surd$ & & $\surd$\\
			Visual feature & & & $\surd$ & $\surd$\\
			\midrule
			Taxi Invoice & 90.34 & 98.45 & 98.71 & \textbf{98.73}\\
			Business Email & 74.51 & 82.88 & 86.02 & \textbf{87.33}\\
			WildReceipt & 72.9 & 87.75 & 83.62 & \textbf{89.62}\\
			Resume & 76.73 & 82.26 & 82.62 & \textbf{83.16}\\
			\bottomrule
		\end{tabular}
	\end{center}
\end{table}

\subsubsection{Effects of different components}  
\begin{figure}[h]
	\centering
	\includegraphics[width=0.45\textwidth]{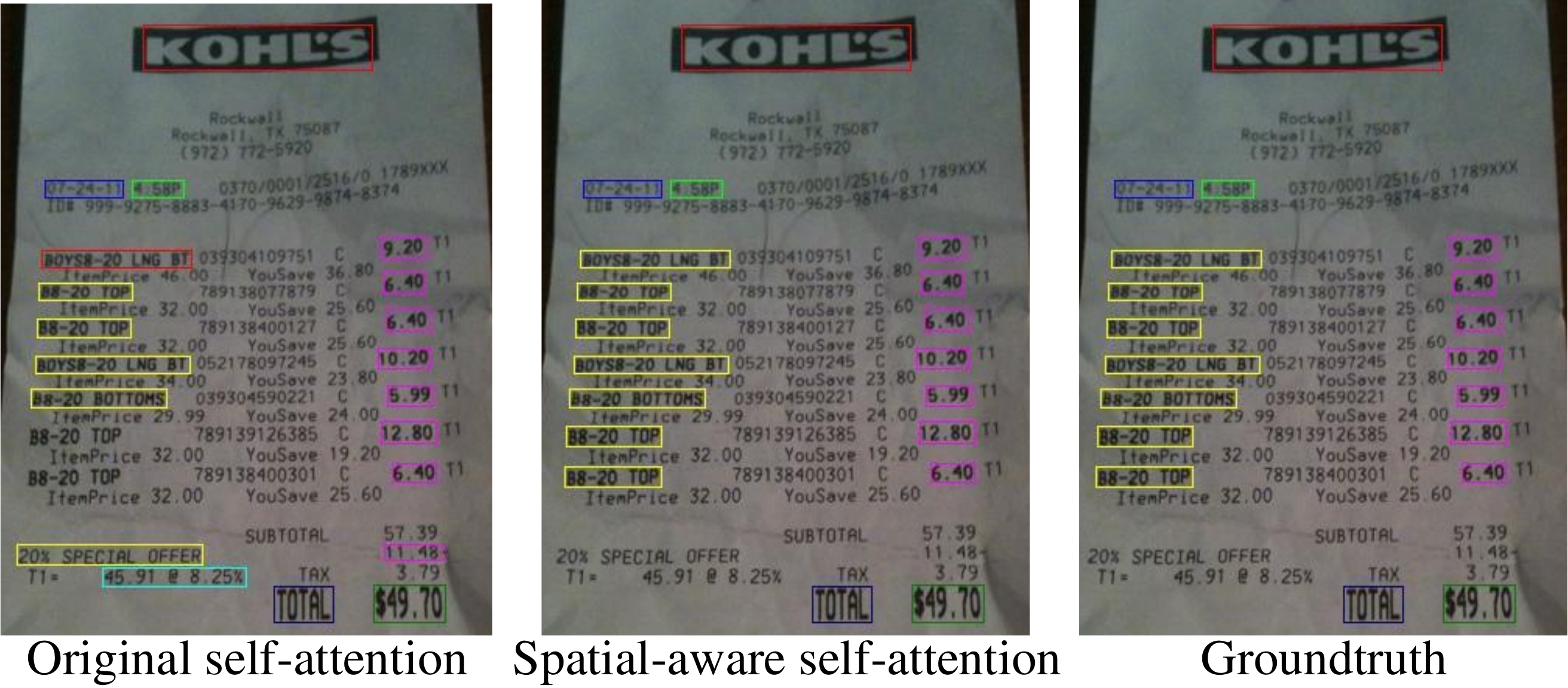}
	\caption{{Visual examples of original self-attention and spatial-aware self-attention.  Different colors denote different entities, such as {\color{red}{Store-Name}}, {\color{blue}{Date}}, {\color{green}{Time}}, {\color{yellow}{Product-Item}}, {\color{cyan}{Product-Quantity}}, {\color{VioletRed}{Product-Price}}, {\color{ForestGreen}{Total}}. Best viewed in color.}}
	\label{fig:ras_vs_sas}
\end{figure}

\begin{figure}[h]
	\centering
	\includegraphics[width=0.49\textwidth]{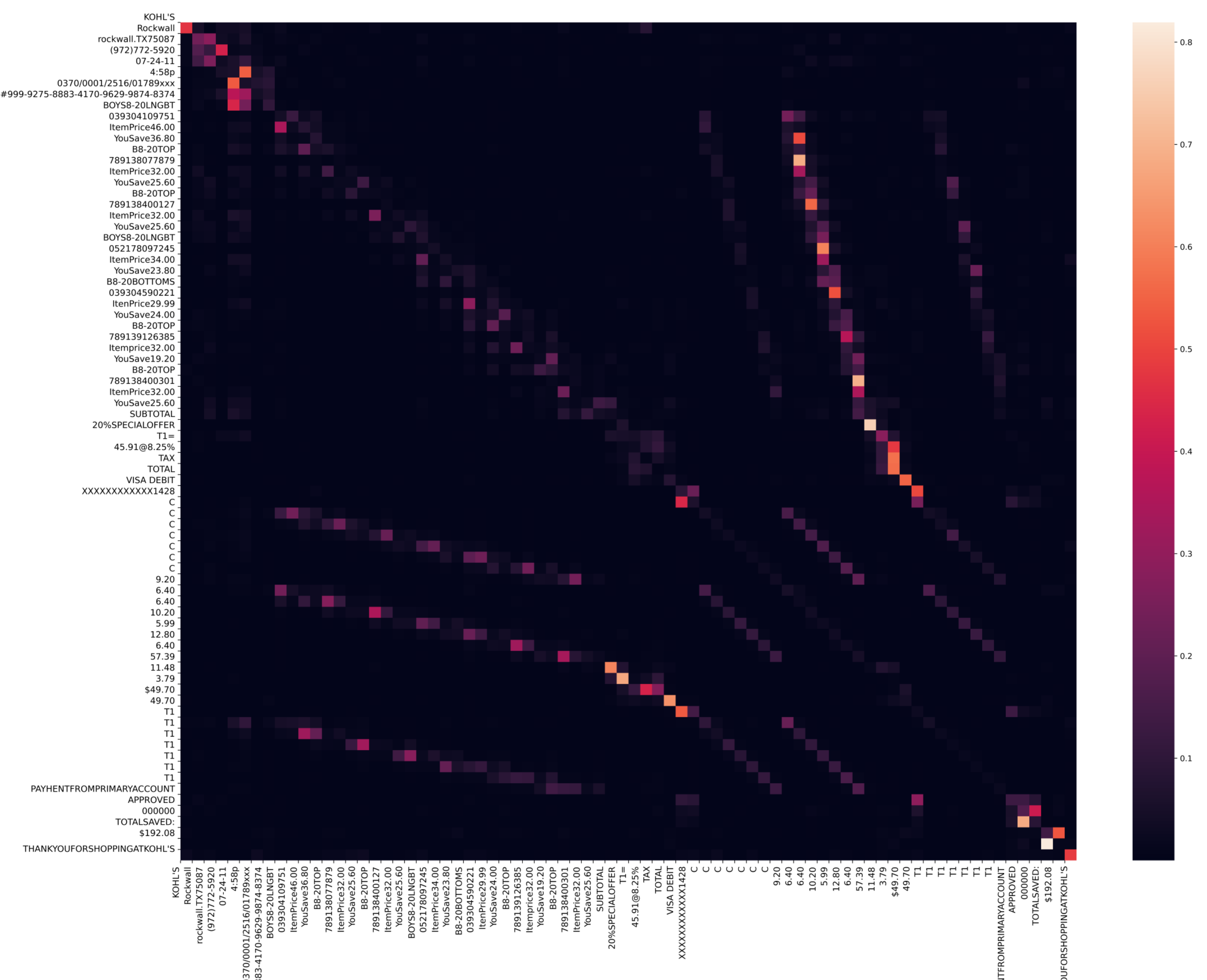}
	\caption{Visualization of spatial-aware self-attention. Total-Key ({\color{yellow}{TOTAL}}) and Total-Value ({\color{yellow}{$\$49.70$}}), Product-Item ({\color{red}{B8-20TOP}}) and Product-Price ({\color{red}{$6.40$}}) always appear together, and their  pairwise relations can be learned. Best viewed in color and zoom in to observe other pairwise relations.}
	\label{fig:vis_sas}
\end{figure}

\zzcedit{
Here we attempt to analyze the effects of the spatial-aware self-attention mechanism,  pre-trained language model, the gating mechanism and Bi-LSTM in IE on four kinds of datasets, as shown in Table \ref{table:components}. 
\begin{table}[h]
	\begin{center}
		\caption{\zzcedit{Accuracy results (eF1) with different components.}}
		\label{table:components}
		\begin{tabular}{lcccc}
			\toprule
			Original self-attention & $\surd$ & & & \\
			Spatial-aware self-attention & & $\surd$ &  $\surd$  &  $\surd$\\
			Language model (Bert) & & &$\surd$  & \\
			language model (LLM v2) & & &  & $\surd$\\
			\midrule
			Taxi Invoice & 98.54 & 98.56 &  \textbf{98.73} & \zzcedit{98.52}\\
			Business Email & 85.25 & 85.73 &  \textbf{87.33}  & \zzcedit{86.14}\\
			WildReceipt & 85.11 & 88.84 &  {89.62}  & \zzcedit{\textbf{90.10}}\\
			Resume & 80.43 & 80.48  & \textbf{83.16}  & \zzcedit{82.57}\\
			\bottomrule
		\end{tabular}
	\end{center}
\end{table}

\textbf{Evaluation of spatial-aware self-attention (short by SaSa).} 
SaSa is devoted to learn layout features to improve the final performance.}
From Table \ref{table:components}, we see that SaSa can boost performance, especially on the WildReceipt.
{This is because, compared to the original self-attention using entities' absolute positions only, the spatial-aware self-attention also makes use of relative position offsets between entities, and learns their pairwise relations. 
Visual examples are shown in Figure.~\ref{fig:ras_vs_sas}.
We see that `Product-Item' and `Product-Price' always appear in pairs.
Spatial-aware self-attention can capture such pairwise relations and then improve model performances. 
Its attention map is visualized in Figure.~\ref{fig:vis_sas}, which demonstrates that the spatial-aware self-attention indeed learns the pairwise relations between entities (\eg  pair of `Total-Key' and `Total-Value', and pair of `Product-Item' and `Product-Price').
}
\begin{figure}[h]
	\centering
	\includegraphics[width=0.45\textwidth]{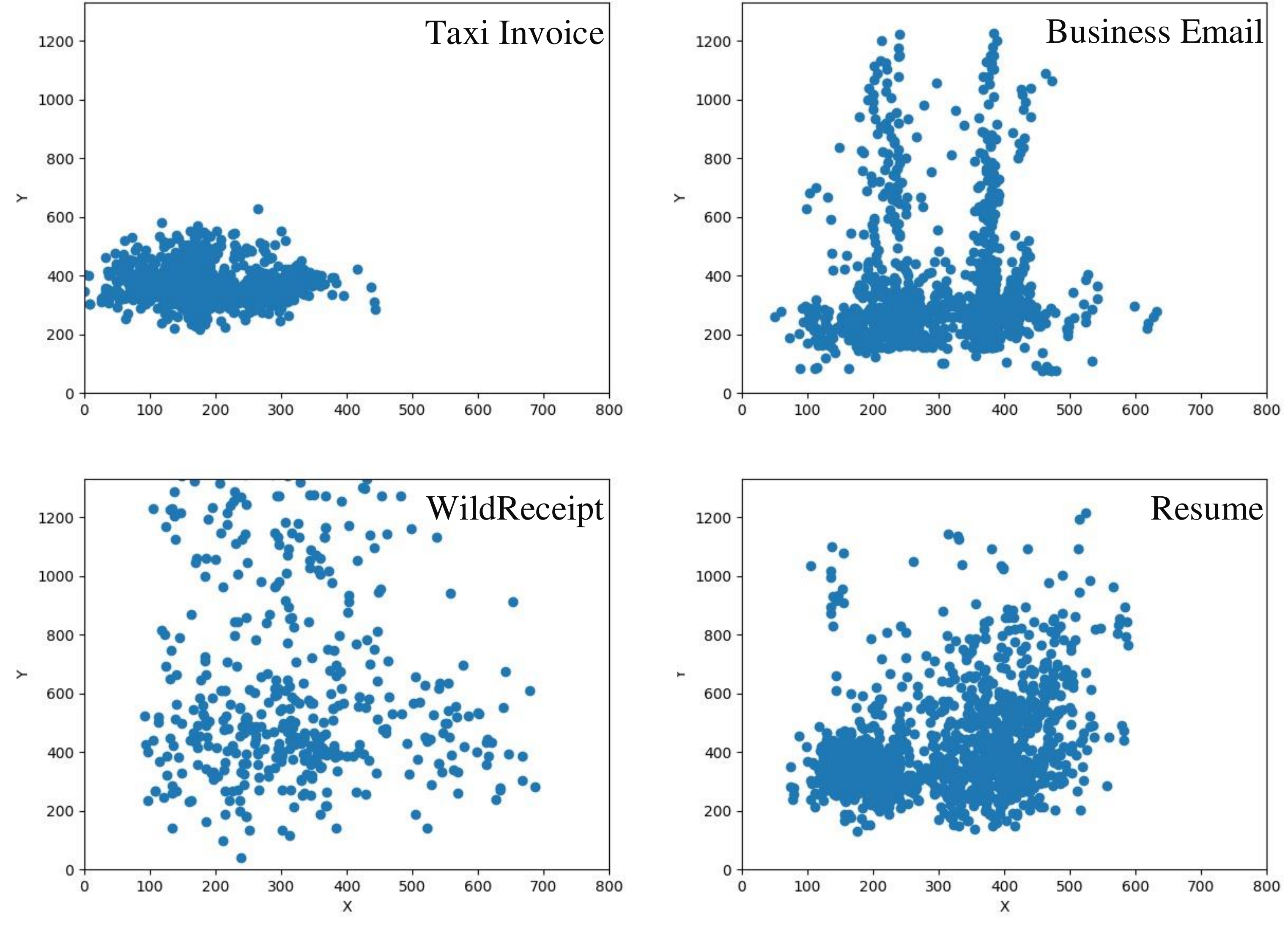}
	\caption{\zzcedit{Layout distribution on four kinds of benchmarks, where positions of entity of `code', `to', `date' and `school' are projected on their canvas.}}
	\label{fig:layout-distri}
\end{figure}
\begin{table}[h]
	\begin{center}
		\caption{\zzcedit{Accuracy results (eF1) by SaSa on more complex scenes.}}
		\label{table:sasa-complex}
		\scalebox{0.95}{
		\begin{tabular}{lcccc}
			\toprule
			Strategy & Taxi Invoice & Business Email & WildReceipt & Resume \\ 
			\midrule
			\emph{wo.} Sasa & 98.45 & 79.93 & 82.28 & 75.96 \\
			\emph{w.} SaSa & \textbf{98.55} & \textbf{83.23} & \textbf{87.63} & \textbf{78.55} \\
			\bottomrule
		\end{tabular}
		}
	\end{center}
\end{table}

\zzcedit{
We also note that improvements by SaSa is not always significant on all benchmarks like Taxi Invoice and Resume. This is because they are dominated with similar layouts (\eg See the layout distribution of Taxi Invoice in Figure. \ref{fig:layout-distri}). 
Thus, effectiveness of SaSa on diversity of layouts has not been measured  well. 
While for WildReceipt with complex layouts, the effects are well demonstrated.

To further demonstrate effects of SaSa, we supplement experiments on the  complex scenes constructed by randomly zooming out the foreground over the background from 1.2x to 3x. Table \ref{table:sasa-complex}  shows the results, which denotes that SaSa can significantly improve the final performance without any side effects.
}

\begin{figure}[t]
	\centering
	\includegraphics[width=0.5\textwidth]{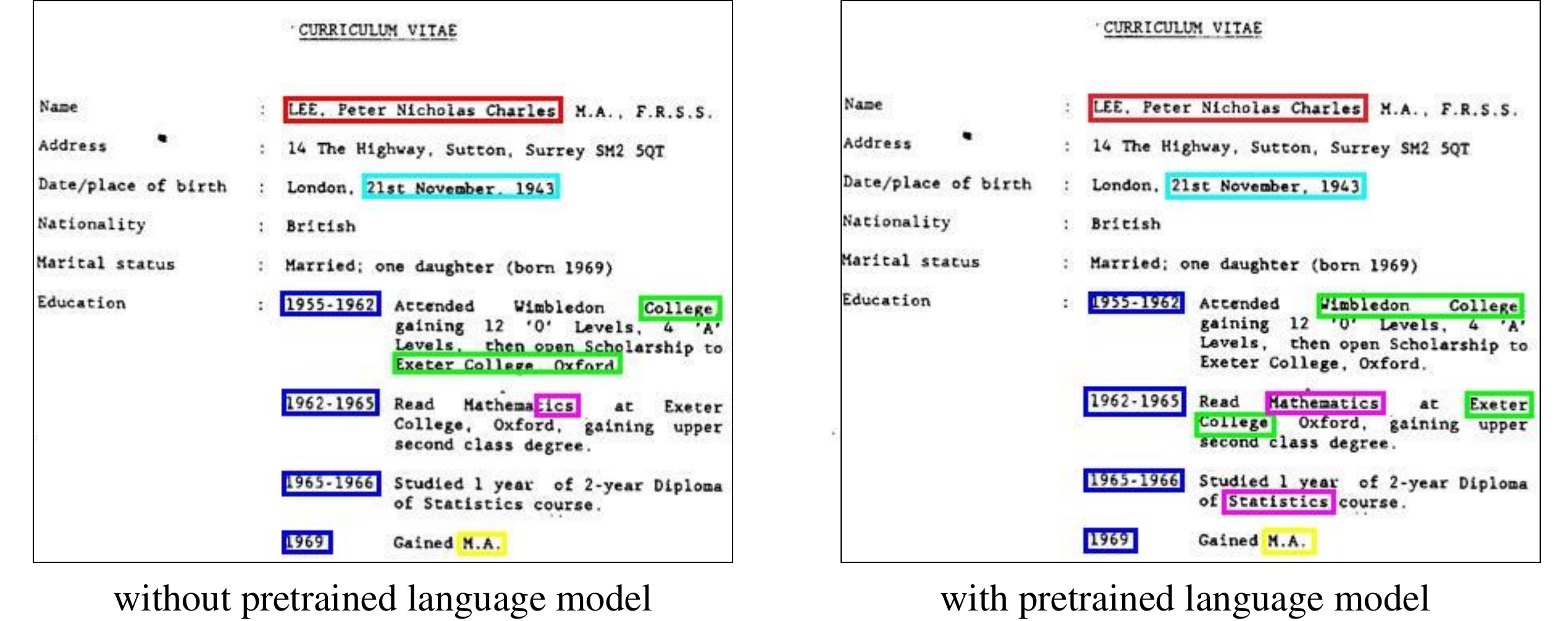}
	\caption{{Illustration of pre-trained language model's effects. Best viewed in color and zoom in.}}
	\label{fig:bert_contri}
\end{figure}

\zzcedit{\textbf{Evaluation of pre-trained language models.} }
When introducing the prior knowledge from Bert \cite{devlin2018bert}, the performance of information extraction is significantly improved on the scenarios that require semantics like WildReceipt, Business Email and Resume. 
{As shown in Figure~\ref{fig:bert_contri}, in the Resume case, introducing the pre-trained language model helps recognize `School' and `Specialty' entities, which are hard to be extracted solely using textual features.
}

\zzcedit{We also attempt to apply LayoutLM v2 as the prior model, showing effective improvements on most of scenes. 
However, it is not always better than results with Bert on the four kinds of documents, as shown in Table \ref{table:components}. 
We think that Bert conveys the pure semantic features while LayoutLM contains both vision and semantic features. 
\zzcedit{The LayoutLM has much overlap to the proposed multi-modal context block, i.e., information redundancy, and bears the problem of double-computing visual features. }
Hence, we use Bert as the pre-trained model by default.
}

\zzcedit{
\textbf{Evaluation of gating mechanism in Equation \ref{gating}.}
We also compare the gating fusion with two classical feature fusion strategies, i.e., the element-wise \emph{Summation} and \emph{Concatenation}. 
The two classical fusion strategies have their strong points on corresponding datasets. 
For example, \emph{Concatenation} operation has better result on Business Email, while falls behind \emph{Summation} on other datasets. 
Benefiting from the on-off role, gating mechanism can help achieve the best performance. 
Table \ref{table:gating} shows the results.
}
\begin{table}[h]
	\begin{center}
		\caption{\zzcedit{Effects (eF1) of gating mechanism.}}
		\label{table:gating}
		\begin{tabular}{lccc}
			\toprule
			Strategy & Concatenation & Summation & Gating \\
			\midrule
			Taxi Invoice & 98.41 & 98.62 & \textbf{98.73} \\
			Business Email & 87.06 & 86.19 & \textbf{87.33}\\
			WildReceipt & 87.95 & 88.47 & \textbf{89.62}\\
			Resume & 81.55 & 82.26  & \textbf{83.16}\\
			\bottomrule
		\end{tabular}
	\end{center}
\end{table}

\zzcedit{\textbf{Evaluation of Bi-LSTM in IE.} }
\zzcedit{
As addressed in Section \ref{ie}, LSTM can help learn relations between characters, which is useful on the token-level scenes like Business Email and Resume. 
Table \ref{table:lstm} shows that LSTM can significantly boost the final performance.
\begin{table}[h]
	\begin{center}
		\caption{\zzcedit{Effects (eF1) of LSTM in IE part.}}
		\label{table:lstm}
		\begin{tabular}{lcc}
			\toprule
			Strategy & Email & Resume \\ 
			\midrule
			\emph{wo.} LSTM & 85.29 & 78.06  \\
			\emph{w.} LSTM & \textbf{87.40} & \textbf{83.24}  \\
			\bottomrule
		\end{tabular}
	\end{center}
\end{table}

}

\subsubsection{Effects of different number of layers and heads} 
Table~\ref{table:performance-layers-heads-global} analyzes the effects of different numbers of layers and heads in the spatial-aware self-attention. 
Taxi Invoices is relatively simple and has a fixed layout. 
Thus the model with 1 or 2 layers and the small number of heads achieves promising results. 
For scenes with complex layout structures like Resumes and WildReceipt, deeper layers and heads can help improve the accuracy results.
In practice, one can adjust these settings according to the complexity of a  task. 
\begin{table}[h]
	\begin{center}
		\caption{Accuracy results (eF1) with different number of layers and heads in spatial-aware self-attention.}
		\label{table:performance-layers-heads-global}
		\begin{tabular}{l|c|ccccc}
			\cline{1-7}
			\multirow{2}{*}{Datasets}      & \multirow{2}{*}{Layers} & \multicolumn{5}{c}{Heads} \\ \cline{3-7}
			&   &2                & 4      & 8      & 16  &32    \\ \cline{1-7}
			\multirow{4}{*}{\makecell[l]{Taxi\\Invoice}} & 1  &  98.27 &  98.57     &  98.45 & 98.62 & 98.00\\
			& 2  &  98.31 & 98.39 & 98.58 & 98.52 & \textbf{98.74} \\
			& 3  &  98.51 & 98.54 & 98.48 & 98.51 & 98.56 \\
			& 4  &  98.44 & 98.58 & 98.41 & 98.70 & 98.59 \\  \cline{1-7}
			\multirow{4}{*}{\makecell[l]{Business\\Email}}       & 1 & 86.05& 86.41& 85.74&  86.94& 86.43 \\
			& 2 & 85.95& 87.51 & 86.78 & 87.33& 87.59 \\
			& 3  & 86.52 & 87.86 & 87.24& 87.15& \textbf{88.01} \\
			& 4 & 86.48& 87.45& 87.82& 87.88&87.64 \\  \cline{1-7}
			\multirow{4}{*}{WildReceipt}       & 1 & 78.17& 87.8& 88.73 & 88.18& 88.67 \\
			& 2 & 86.26 & 88.11 & 88.21& 89.16& 89.11  \\
			& 3  & 77.1 & 88.62 & 88.95 & 89.48& 89.69 \\
			& 4 & 85.48 & 89.00 & 88.63 & 89.66 &\textbf{90.15}  \\  \cline{1-7}
			\multirow{4}{*}{Resume}       & 1 & 82.18& 82.52 & 81.99 & 81.83&82.49 \\
			& 2 & 82.7 & 82.56 & 82.97 & 82.83&83.57 \\
			& 3  & 82.86& 82.09& 83.05& 82.78&82.96   \\
			& 4 & 82.75 & 83.12 & 82.43 & 82.98&\textbf{83.46} \\  \cline{1-7}
		\end{tabular}
	\end{center}
\end{table}
\begin{table}[h]
	\begin{center}
		\caption{\zzcedit{Comparisons of pipelines and end-to-end training framework. 
		}}
		\label{table:performance-e2e-vs-pipeline}
		\begin{tabular}{c|c|c|c|c}
			 \cline{1-5}
			{Dataset} & {Method} & {\makecell{Detection\\(F$_d$-m)}} & {\makecell{Spotting\\(F$_r$-m)}} & \makecell{IE\\(eF1)}\\ 
			\cline{1-5}
			\multirow{3}{*}{Taxi Invoice} 
			            & base1 & \textbf{95.72}  &\textbf{91.15}  & 88.29 \\
			            & base2 & 95.21  &91.05  & 88.28 \\
			            & e2e & 94.85  &91.07  & \textbf{88.46} \\
			\cdashline{1-5}[1pt/1pt]
			\multirow{3}{*}{Business Email} 
			            & base1 & 97.12  &55.88  & 45.24 \\
			            & base2 & 97.10  &56.18  & 45.47 \\
			            & e2e & \textbf{97.22}  &\textbf{56.83} &  \textbf{45.71} \\
			 \cdashline{1-5}[1pt/1pt]
			\multirow{3}{*}{WildReceipt} 
			            & base1 & 90.31  &73.52  & 69.37 \\
			            & base2 & 90.55  &74.98  & 71.15 \\
			            & e2e & \textbf{90.73}  &\textbf{76.50} &  \textbf{73.12} \\
			 \cdashline{1-5}[1pt/1pt]
			\multirow{3}{*}{Resume} 
			            & base1 & 96.71  &55.15  & 58.53 \\
			            & base2 & 96.86 & 55.56  & 58.31 \\
			            & e2e & \textbf{96.88}  &\textbf{55.66} &  \textbf{58.77} \\
			\cline{1-5}
		\end{tabular}
	\end{center}
\end{table}
\begin{table}[h]
	\begin{center}
		\caption{\zzcedit{The relative $eF1$ for IE \emph{w.r.t} $F_r\text{-}m$ (cared text instances) of text spotting.}}
		\label{table:rir}
		\begin{tabular}{lcccc}
			\toprule
			Method & \makecell{Taxi\\ Invoice} & \makecell{Business\\ Email} & WildReceipt & Resume \\
			\midrule
			base2 & 99.41 & \textbf{95.76} & 94.68 & 95.12\\
			e2e & \textbf{99.45} & 95.01 & \textbf{96.11} & \textbf{97.41}\\
			\bottomrule
		\end{tabular}
	\end{center}
\end{table}

\begin{table*} [t]
\caption{\zzcedit{Comparison (\emph{eF1}) with the state-of-the-arts on FUNSD, SROIE, EPHOIE, and WildReceipt. 
$^*$ refers to the results are reported by our re-implemented model. 
On SROIE dataset, \textit{Italic} denotes the segment-level accuracy while others are token-level accuracy.
}
}
\label{table:sotas}
\begin{center}
\scalebox{0.93}{
\begin{tabular}{l|ccc|ccc|ccc|ccc|c}
\cline{1-14}
\multirow{2}{*}{Methods} & \multicolumn{3}{c|}{FUNSD}  & \multicolumn{3}{c|}{SROIE} & \multicolumn{3}{c|}{EPHOIE} & \multicolumn{3}{c|}{WildReceipt} & Speed\\
\cline{2-14}
 & $eREC$ & $ePRE$  & $eF1$ & $eREC$ & $ePRE$  & $eF1$ & $eREC$ & $ePRE$  & $eF1$ & $eREC$ & $ePRE$  & $eF1$ & \zzcedit{\emph{Avg} FPS} \\
\cline{1-14}
	LLM\cite{xu2019layoutlm}	&82.19   & 75.96  & 78.95  & 95.24 & 95.24 & 95.24  & - &- & - & -& - & - & -\\
	LLMv2\cite{Xu2020LayoutLMv2MP}	&85.19   & 83.24  &  84.20 & 96.61 & 96.61 & 96.61  & \zzcedit{99.51$^*$} & \zzcedit{99.06$^*$} & \zzcedit{\textbf{99.28}$^*$} & \zzcedit{91.78$^*$}& \zzcedit{92.45$^*$} & \zzcedit{\textbf{92.05}$^*$}  & 3.93 \\
	SLM\cite{li2021structurallm}	&86.81   &  83.52 &  \textbf{85.14} & - & - & -  & - &- & - & -& - & - & -\\ 
	StrucText\cite{li2021structext}	&80.97   & 85.68  &  83.09 & \tabincell{c}{{98.81} \\ \textit{98.52}} & \tabincell{c}{92.77 \\ \textit{95.84}} & \tabincell{c}{95.62 \\ \textit{96.88}}  & - &- & 97.95 & -& - & - & -\\
	LAMBERT\cite{garncarek2021lambert}	&-   & -  &  - &- & - & \textbf{98.17}  & - &- & - & -& - & - & -\\
	TILT\cite{powalski2021going}	&-   & -  &  - &- & - & 98.10  & - &- & - & -& - & - & -\\
	SelfDoc\cite{li2021selfdoc}	&-   & -  &  83.36 &- & - & -  & - &- & - & -& - & - & -\\
	\cline{1-14}
	SPADE\cite{hwang2020spatial}	&-   &  - &  70.50 & - & - &  - & - &- & - & -& - & - & -\\
	LSTM-CRF\cite{lample2016neural}	&-   &  - & 62.13  & - & - & 90.85  & - &- & 89.10 & 82.60$^*$& 83.90$^*$ & 83.20$^*$ & 5.43 \\
	Chargrid\cite{katti2018chargrid}	&39.98$^*$   &  73.45$^*$ & 50.50$^*$  & - & - & 80.9  & 77.82$^*$ &73.41$^*$ & 75.23$^*$ & 75.64$^*$& 75.22$^*$ & 75.39$^*$  & 12.42 \\
	GAT\cite{velivckovic2018graph}	&70.81$^*$   &  71.03$^*$ &  70.73$^*$ & 83.14$^*$ & 91.73$^*$ &  87.23$^*$ & 97.36$^*$ &96.48$^*$ & 96.90$^*$ & 84.57$^*$& 86.37$^*$ & 85.43$^*$  & \textbf{15.73} \\
	VRD\cite{liu2019graph}	&-   &  - &  72.43 & - & - &  95.10 & - &- & 92.55 & 84.57& 86.37 & 85.70 & -\\
	GraphIE\cite{qian2019graphie}	&-   & -  &  72.12 & - & - & 94.46  & - &- & 90.26 & -& - & - & -\\
	VIES\cite{wang2021towards}	&-   & -  &-   & - & - & 96.12  & - &- & 95.23 & -& - & - & -\\
	PICK\cite{PICK2020YU}	&-   & -  & -  & 95.46 & 96.79 &  96.12 & - &- & - & -& - & - & -\\
	MatchVIE\cite{tang2021matchvie}	&-   & -  & 81.33  & - & - &  96.57 &  -&- & 96.87 & -& - & - & -\\
	TextLattice\cite{wang2021tag}	&-   &  - & -  &  -& - &  96.54 & - &- & 98.06 & -& - & - & -\\
	SDMG-R\cite{sun2021spatial}	&-   & -  & -  & - & - &  87.10 & - &- & - & -& - & 88.70 & -\\ 
	\cdashline{1-14}
	\zzcedit{TRIE\cite{zhang2020trie} \emph{w. gt}}	&-   &  - & 78.86 & - & - & 96.18  & - &- & 93.21 & 84.69& 87.58 & 85.99 & 14.52\\
	\zzcedit{TRIE++ \emph{w. gt}}	&\textbf{82.94}   &  \textbf{84.37} & \textbf{83.53}  & \tabincell{c}{\textbf{95.89} \\ \textbf{\textit{98.40}}} & \tabincell{c}{\textbf{97.72} \\ \textbf{\textit{98.35}}} & \tabincell{c}{\textbf{96.80} \\ \textit{\textbf{98.37}}}  & \textbf{98.96} &\textbf{98.75} & \textbf{98.85} & \textbf{89.59}& \textbf{90.84} & \textbf{90.15} & 13.88\\
        \cline{1-14}
\end{tabular}
}
\end{center}
\end{table*}

\begin{table*} [t]
\caption{\zzcedit{Strong baselines for detection, end-to-end text spotting and information extraction on four kinds of datasets, i.e., Taxi Invoice, Business Email, WildReceipt and Resume. Their entity-level results are shown at  \emph{https://davar-lab.github.io/publication/trie++.html.} 
`TR' means the text reading module in \emph{base2} (See Table \ref{table:performance-e2e-vs-pipeline}) is used for end-to-end evaluation.
}
}
\label{table:baseline}
\begin{center}
\scalebox{1}{
\begin{tabular}{l|l|c|c|c|c|c|c|c|c|c|c}
\cline{1-12}
\multirow{2}{*}{Dataset} & \multirow{2}{*}{Methods} & \multicolumn{3}{c|}{Text Detection}  & \multicolumn{3}{c|}{Text Spotting} & \multicolumn{3}{c|}{End-to-end IE} & Speed\\
\cline{3-12}
 & & $REC_d$ & $PRE_d$  & $F_d\text{-}m$ & $REC_r$ & $PRE_r$  & $F_r\text{-}m$ & $eREC$ & $ePRE$  & $eF1$ & $FPS$\\
\cline{1-12}
\multirow{10}{*}{\makecell{Taxi\\invoice}}       
        & PPOCR+IE$_{\text{TRIE++}}$   &  76.73 &  59.13 & 66.79 & 50.23 &  38.72 & 43.73 & 49.66 & 52.85 & 50.87 & {4.78}\\
        & Tesseract+IE$_{\text{TRIE++}}$   & 21.96  &  45.20 & 29.56 & 2.91 & 5.99  & 3.92 & 6.98 & 18.24  & 9.68 & 1.53\\
        \cdashline{2-12}[1pt/1pt]
		& TR+Chargrid  &  \multirow{5}{*}{99.18}  & \multirow{5}{*}{91.55}  & \multirow{5}{*}{95.21}  &  \multirow{5}{*}{94.85}  & \multirow{5}{*}{87.55}  & \multirow{5}{*}{91.05}  & 92.11 & 84.6 & 88.14& 2.83\\
		& TR+GAT  &   &   &  &  &   &  & 91.58 & 84.35 &  87.78 & 3.86\\ 
		& TR+LSTM-CRF   &   &   &  &  &   &  &90.73 & 83.2 & 86.74& 3.62\\
		& TR+Bert-CRF   &   &   &  &  &  &  & 88.24 & 85.25  & 86.74 & {3.25} \\
		& TR+LLM v2   &   &   &  &  &  &  & 88.91 & 85.42  & 87.14 & 2.86 \\
		& TR+IE$_{\text{TRIE++}}$  & 99.18  & 91.55  & \textbf{95.21} & 94.85 & 87.55 & 91.05 &92.15 & 84.8 & 88.28& 3.69\\ 
		\cdashline{2-12}[1pt/1pt]
		& TRIE   &  99.18 &  91.45 & 95.16 & 94.78 & 87.39 & 90.94 & 91.95 & 84.74 & 88.16 & \textbf{5.10} \\
		&TRIE++   & 99.22 & 90.85 & 94.85 &95.27  & 87.23  & \textbf{91.07}  & 92.5 & 84.85& \textbf{88.46}& {4.26}\\ 
        \cline{1-12}
		\cline{1-12}
\multirow{10}{*}{\makecell{Business\\Email}}   
        & PPOCR+IE$_{\text{TRIE++}}$   & 89.79  &  82.92 & 86.22 & 11.30 &  10.43 & 10.85 & 8.45& 8.74 & 8.45 & {2.55}\\
        & Tesseract+IE$_{\text{TRIE++}}$   & 68.96  & 81.07  & 74.53 & 25.25 & 29.68  & 27.28 & 23.90 & 27.39  & 25.39  & 1.73\\
        \cdashline{2-12}[1pt/1pt]
		& TR+Chargrid   &  \multirow{5}{*}{96.4}  & \multirow{5}{*}{97.81}  & \multirow{5}{*}{97.10}  &  \multirow{5}{*}{55.77}  & \multirow{5}{*}{56.59}  & \multirow{5}{*}{56.18}  &32.18 & 33.27 &32.64 & {2.30}\\
		& TR+GAT &   &   &  & & & & 42.34  & 44.22& 43.09 & {2.30}\\ 
		& TR+LSTM-CRF   &   &   &  & & & &43.27 & 45.43 &44.2 & 1.30\\
		& TR+Bert-CRF   &   &   &  &  &  &  & 42.62 & 47.84  & 44.76 & {1.80} \\
		& TR+LLM v2   &   &   &  &  &  &  & 45.22 & 48.49  & \textbf{46.68} & 1.50 \\
		& TR+IE$_{\text{TRIE++}}$  &  96.4 &  97.81 & 97.10 & 55.77 & 56.59 & 56.18 &44.92 & 46.32 & 45.47& 2.22\\
		\cdashline{2-12}[1pt/1pt]
		& TRIE   &  96.66 &  97.72 & 97.18 & 56.02 & 56.64 & 56.33 & 40.43 & 44.96 & 42.01 & \textbf{2.65} \\
		&TRIE++  & 96.62 & 97.83 & \textbf{97.22} & 56.48  & 57.18  & \textbf{56.83}  & 45.7  & 45.85& {45.71} & {2.30}\\ 
     \cline{1-12}
		\cline{1-12}
\multirow{10}{*}{\makecell{WildReceipt}}        
        & PPOCR+IE$_{\text{TRIE++}}$   & 70.32  &  77.42 & 73.70 & 33.74 & 37.14  & 35.36 & 33.91 & 40.14 & 35.73 & {3.75}\\
        & Tesseract+IE$_{\text{TRIE++}}$   &  41.11 &  58.25 & 48.20 & 12.20 & 17.29  & 14.31 & 11.14 & 21.99  & 14.59  & 1.26\\
        \cdashline{2-12}[1pt/1pt]
		& TR+Chargrid   &  \multirow{5}{*}{90.65}  & \multirow{5}{*}{90.45}  & \multirow{5}{*}{90.55}  &  \multirow{5}{*}{75.06}  & \multirow{5}{*}{74.89}  & \multirow{5}{*}{74.98}  &61.27 & 60.92 & 61.04& 3.97\\
		& TR+GAT &   &   &  &  &   &  & 68.16 & 69.21 &  68.66 & 4.25\\ 
		& TR+LSTM-CRF   &   &   &  &  &   &  &65.85 & 67.95 & 66.84& 2.81\\
		& TR+Bert-CRF   &   &   &  &  &  &  & 69.21 & 72.24  & 70.71 & {3.26} \\
		& TR+LLM v2   &   &   &  &  &  &  & 72.49 &  74.07 & \textbf{73.26} & 2.31 \\
		& TR++IE$_{\text{TRIE++}}$   & 90.65 & 90.45  & 90.55 & 75.06 & 74.89 & 74.98 &71.05 & 71.32 & 71.15& 4.10\\
		\cdashline{2-12}[1pt/1pt]
		& TRIE   &  90.42 &  90.79 & 90.61 & 75.29 & 75.60 & 75.44 & 70.24 & 66.81 & 68.19 & \textbf{6.21} \\
		&TRIE++  & 90.23 & 91.22 & \textbf{90.73} & 76.09  & 76.92 & \textbf{76.50}  & 72.33 &  74.04& {73.12} & \textbf{6.21}\\ 
     \cline{1-12}
		\cline{1-12}
\multirow{10}{*}{{Resume}}        
        & PPOCR+IE$_{\text{TRIE++}}$   &  92.18 &  85.67 & 88.81 & 22.49 & 20.90  & 21.66 & 26.65 & 26.41 & 26.51 & {1.27}\\
        & Tesseract+IE$_{\text{TRIE++}}$   &  60.35 & 80.17  & 68.86 & 23.46 &  31.16 & 26.76 & 27.48 &  28.56 & 27.99  & 1.20\\
        \cdashline{2-12}[1pt/1pt]
		& TR+Chargrid   &  \multirow{5}{*}{96.03}  & \multirow{5}{*}{97.71}  & \multirow{5}{*}{96.86}  &  \multirow{5}{*}{55.09}  & \multirow{5}{*}{56.05}  & \multirow{5}{*}{55.56}  &41.46 & 38.18 &39.62 & 2.10\\
		& TR+GAT &   &   &  &  &   &  & 55.96  & 53.42& 54.65 & 2.20\\ 
		& TR+LSTM-CRF   &   &   &  &  &   &  &55.63 & 54.02 &54.64 & 0.71\\
		& TR+Bert-CRF   &   &   &  &  &  &  & 56.96 & 57.67  & 57.31 & {1.43} \\
		& TR+LLM v2   &   &   &  &  &  &  & 57.33 & 57.70 & 57.49 & 0.90 \\
		& TR+IE$_{\text{TRIE++}}$   & 96.03  &  97.71 & 96.86 & 55.09 & 56.05 & 55.56 &59.08 & 57.62 & 58.33& 2.09\\
		\cdashline{2-12}[1pt/1pt]
		& TRIE    &  95.99 &  97.82 & \textbf{96.89} & 55.16 & 56.21 & \textbf{55.68} & 55.94 & 56.67 & 56.29 & \textbf{2.39} \\
		&TRIE++& 95.96 &  97.82 & {96.88} & 55.13  & 56.20  & {55.66}  & 59.43  & 58.16& \textbf{58.77}& {2.31}\\ 
        \cline{1-12}
\end{tabular}
}
\end{center}
\end{table*}

\subsubsection{Effects of the end-to-end training}  
To verify the effects of the end-to-end framework on text reading {and information extraction}, we perform the following experiments on four kinds of VRD datasets.
{
We first define two strong baselines for comparison. 
(1) \textit{Base1}. The detection, recognition and information extraction modules are separately trained, and then pipelined as an inference model.
(2) \textit{Base2}. The detection and recognition tasks are jointly optimized, and then pipelined with the separately trained information extraction task.
While joint training of the three modules is denoted as our \textit{end-to-end} \zzcedit{(short by \emph{e2e})} framework.
}
Notice that all multi-modal features (See Section \ref{forward_effect}) are integrated. 
The layer and head numbers in self-attention are set as (2, 2, 4, 2) and (32, 32, 16, 32) for four different tasks (Taxi Invoice, Business Email, WildReceipt, Resume in order), respectively.
\zzcedit{
Results are as shown in Table~\ref{table:performance-e2e-vs-pipeline}, in which the end-to-end training framework can benefit both {text reading} and {information extraction} tasks. 

Concretely, 
on \textit{text reading}, \textit{e2e}  surpasses \textit{base1} and \textit{base2} on most scenes (WildReceipt, Business Email and Resume), while slightly falls behind base1 by 0.08 on Taxi Invoice dataset. 
In fact, the final results on \textit{information extraction} are important. 
We see that \textit{e2e} obtains the best performance on all datasets, achieving the largest eF1 performance gain (3.75\%) on WildReceipt. 
We also note that, on Taxi Invoice and Resume cases, improvements by \textit{e2e} are not significant. 
There are two main reasons. 
(1) The IE part has achieved the saturated performance \textit{w.r.t} the cared text in text spotting. 
And the slight performance gain on IE is hard-won. 
Therefore, we calculate the relative \emph{eF1}  of IE \textit{w.r.t} the cared text in text spotting (i.e., $\frac{eF1~ \text{of}~ \text{IE}}{\text{cared}~ F_r\text{-}m~ \text{of}~ \text{spotting}}$) to  obtain the relative IE performance, which gets rid of the impacts of text spotting.
Table \ref{table:rir} shows the results. 
For Taxi Invoice, the performance is approximate to the upper-bound (100\%), the improvement is limited. And for WildReceipt and Resume, the improvements are significant.
As to Business Email, though \emph{base2} has better result (Table \ref{table:rir}) on the evaluation of relative performance, \emph{e2e} boosts the final results in Table \ref{table:performance-e2e-vs-pipeline}. 
(2) The text spotting capability limits the final results directly. 
It means that  text spotting maybe the real performance bottleneck for  end-to-end VRD information extraction. 
However, Taxi Invoice contains many low-quality (\eg blur or occlusion text) text, while Resume has amount of long text instance (once one character is  recognized wrongly, the final results is wrong.). Techniques for handling  the above obstinate problems are needed. 

In sum, compared to the strong baselines, the end-to-end trainable strategy is actually working, and then boost the final performance. 

}
\subsection{Comparisons with the State-of-the-Arts}
\label{sota}
Recent methods \cite{xu2019layoutlm,Xu2020LayoutLMv2MP,li2021structurallm,li2021structext} focused on the information extraction task by adding great number of extra training samples like IIT-CDIP dataset \cite{Lewis2006BuildingAT} and DocBank~\cite{li2020docbank}, and then have impressive results on the downstream datasets. 
Following the typical routine, we also compare our method with them on several popular benchmarks. 
\zzcedit{Note that, all results are reported by using the official OCR annotations.}

\textbf{Evaluation on FUNSD} 
The dataset is a noisy scanned from the dataset with 200 images. The results are shown in FUNSD column of Table~\ref{table:sotas}. 
To be fair, we first compare our method with those without introducing extra data. 
Our method significantly outperforms them with a large margin (83.53 \emph{v.s.} 81.33 of MatchVIE\cite{tang2021matchvie}).
When comparing with models trained with extra data, our method is still competitive. 
It only falls behind the LLMv2\cite{Xu2020LayoutLMv2MP} and SLM\cite{li2021structurallm}. 

\textbf{Evaluation on SROIE} 
The dataset has 963 scanned receipt images, which is evaluated on four entities in many works. 
Most of the results are impressive, as shown in SROIE column of Table~\ref{table:sotas}. This is because methods tend to achieve the performance upper bound of this dataset. For example, StrucText \cite{li2021structext} (with extra data) has achieved 96.88 of \emph{eF1}, which only has slight advantage over 96.57 of MatchVIE\cite{tang2021matchvie}. 
{Our method shows promising results on this benchmark, with 96.80 $eF1$ in the token granularity (same to most works~\cite{PICK2020YU,wang2021tag,wang2021towards,tang2021matchvie,xu2019layoutlm,Xu2020LayoutLMv2MP,zhang2020trie}) and 98.37 in the segment granularity (same to StrucText~\cite{li2021structext}).}

\textbf{Evaluation on EPHOIE} 
The dataset is a Chinese examination paper head dataset. 
Our method obviously surpasses previous methods \zzcedit{without extra data, achieving the best result (98.85\%), and is competitive to the large model LLMv2 \cite{Xu2020LayoutLMv2MP}. }
Similar to SROIE, its performance upper bound is limited. 
That is, only 1.15\% of improvement space is left.

\textbf{Evaluation on WildReceipt} 
This receipt dataset \cite{sun2021spatial} is more challenging than SROIE, which is collected  from  document images  with  unseen  templates  in  the  wild. 
Most of the methods like  GAT\cite{velivckovic2018graph} have  rapid performance degradation compared to results in SROIE and EPHOIE. 
While our method still has the best result (90.15\% of \emph{eF1}) compared to existing methods \zzcedit{trained without extra data}, which verifies the advantages of the proposed method.

\zzcedit{
\textbf{Inference Speed}. 
We test the inference speed (the average frames per second, \emph{avg} FPS) of the re-implemented models for evaluating their efficiency. 
Our model surpass LLMv2 \cite{Xu2020LayoutLMv2MP} largely, and has competitive efficiency with Chargrid \cite{katti2018chargrid} and GAT \cite{velivckovic2018graph}. 

In sum, compared to methods trained without extra data, the proposed method achieves the new state-of-the-arts on the pure information extraction task, and also shows strong generalization ability across various VRDs. 
Even compared with the large models, our method is also competitive.
}

\subsection{Strong Baselines on Four Categories of VRD} \label{baseline}
For the pure information extraction task, their results (as shown in Table \ref{table:sotas}) are calculated based on the ground truth of detection and recognition. 
However, the influence of OCR should not be neglected in reality.
Considering the real applications, i.e., \zzcedit{end-to-end extracting information from VRD}, one way is to divide the task as two pipelined steps: (1) obtaining text spotting results with a public OCR engines, (2) and then performing the information extraction. 
We here provide \zzcedit{several groups of solid baselines} on four kinds of VRDs. 

\zzcedit{
\subsubsection{Comparisons among Different Experimental Settings} 
We first build the basic baselines by combining the public OCR engines (including PPOCR \cite{du2020pp} and Tesseract \cite{smith2007overview}) and the pre-trained information extraction model (denoted as IE$_{\text{TRIE}}$) used in Section \ref{sota}.
%
Furthermore, we use the \textit{text reading} (denoted as TR) part in base2 (See Table \ref{table:performance-e2e-vs-pipeline}) as the OCR engine, and take the results as the input of information extraction counterparts, i.e., GAT \cite{velivckovic2018graph}, LSTM-CRF \cite{lample2016neural,ma2019end}, Chargrid \cite{katti2018chargrid}, Bert-CRF \cite{devlin2018bert} and LayoutLM \cite{Xu2020LayoutLMv2MP}. 
Concretely, LSTM-CRF \cite{ma2019end} is a classic 1D information extraction paradigm. 
Its input embedding and hidden units in the following BiLSTM are all set to $128$, followed by a CRF layer. 
Chargrid \cite{katti2018chargrid} is a 2D-CNN information extraction framework. The input character embedding is set to $128$  and the rest of network is identical to the paper.
As for graph-based information extraction framework, we adopt GAT\cite{velivckovic2018graph}, which is similar to \cite{liu2019graph}. 
Bert-based and LayoutLM-based model are treated as the large model for information extraction. 
As a result, five kinds of strong baselines are constructed. They are `TR+GAT', `TR+LSTM-CRF', `TR+Chargrid', `TR+Bert-CRF' and `TR+LLM v2'. 
Results are shown in Table \ref{table:baseline}. 
}



\zzcedit{
\textbf{Comparison with Public OCR Engines}. As expected, it is irresponsible to apply open OCR engines on the these specific scenes directly, showing poor results on text spotting and information extraction. 

\textbf{Comparison with Classical Methods}. We compare our method with several classic methods (including `TR+GAT', `TR+LSTM-CRF' and `TR+Chargrid'), and our method obviously outperforms them on all scenes. 

\textbf{Comparison with Large Models}. Since large models (\eg Bert or LayoutLM) can boost the IE performance attributing to its big model of capacity as well as amount of samples, we here also construct such experimental settings for comparison.
We see that our method is superior to the Bert-based method, and achieves competitive results compared to LayoutLM v2-based method, i.e., only falling behind it on the WildReceipt and Business Email scenes.  

}

\subsubsection{Comparison of Inference Speed} 
We evaluate the running time of our model and its counterparts in frames per second (\emph{abbr}. FPS).
Results are as shown in the last column of  Table~\ref{table:baseline}.
{Thanks to feature sharing between \textit{text reading} and \textit{information extraction} modules, \zzcedit{our end-to-end framework runs faster than its pipeline counterparts, especially compared to the large models (\eg LayoutLM v2).
Because of the enhanced multi-modal context block, the proposed method is slightly slower than its conference version (TRIE \cite{zhang2020trie})
}
A more prominent trend is that the algorithm runs faster in scenarios where the length of texts is short in a document (\eg Taxi Invoice and WildReceipt), while on Resume/Business Email datasets with long texts, the FPS drops slightly.
}

\subsubsection{Evaluations among Different Modules}
In the detection part, all methods achieve the satisfactory performance of \emph{F$_d$-m} (larger than 90\%), while the performance on WildReceipt is the lowest. 
This is because the receipt images in WildReceipt are captured in the wild, and they are of non-front views, even with folds. 
When considering the end-to-end text spotting task, results on Business and Resume are poor due to the problems of character distortion and long text. 
{This problem will be a new research direction for OCR.}
For the end-to-end information extraction, results on Business Email are the worst, and the second-worst is Resume. 
{It reveals that there is plenty of work to do concerning end-to-end information extraction.} 

From the perspective of systems, we surprisingly discover that the text recognition may be the top bottleneck for end-to-end understanding VRD on Category II, III and IV. 
The information extraction is another bottleneck due to the complex layouts and long character sentence (Referring to Table \ref{table:baseline}, \ref{table:performance-contribution} and \ref{table:components}). 
Luckily, the end-to-end training strategy can enhance both the text reading and the final information extraction task.
In future, more attention should be paid to the effects of text reading \emph{w.r.t} information extraction.

{
\zzcedit{
\section{Discussion}
\subsection{Effects of Large Pre-trained Model}
The large pre-trained model (short by LPM) becomes a hot research topic in many research communities, while discussions on its industrial application are rarely reported. 

From our point of view, LPM can help achieve better baseline performance, and has good generalization on various scenarios. It is suited to platform applications like AI Open Platform, providing basic trials for various applications. 
However, it is not `green' to training or maintaining LPM, where `green' means lightweight and electric economic. 
LPM also suffers from the computational cost problem, which is impractical when applied in terminal devices (\eg receipt scanner used in logistics industry). 
Perhaps it is a practical problem on how transfer beneficial knowledge from a large model to a small model.
In fact, when applied model on the specific scene, finetuning on the corresponding  data is  essential for better results. 
In this way, performance gain from LPM may be limited. 
Unlike them, our method belongs to another researching routine.


\subsection{Necessity of End-to-End Model}
We here point out that the public OCR engines like PaddleOCR or Tesseract are irresponsible to apply on the vertical commerce scenes directly, as shown in Table \ref{table:baseline}. 
Therefore, it is inevitable to train/finetune OCR models on the corresponding scenarios. 
End-to-end framework can help achieve the global optimization, and also reduce model maintenance cost.

\subsection{Annotation Cost Problem}
Annotation cost is a realistic problem.  
Unfortunately, it is hard to be avoided in real commerce applications (\eg financial auditing). 
Once users (party A) pay money for applications, they tend to have high accuracy requirements but with few samples. 
For deep model, better application performance means more data or annotations. 
Developers (party B) have to annotate labels as detailed  as possible on the given datasets for better results. 
Then some researchers attempt to pre-train large model with big data (\eg collected from internet or other public ways) to relieve the problem. 
However, once when one applies model on specific scenes, developers have to finetune models on the given dataset for better performance. 
In fact, previous methods (\eg layoutLM) also relied on the fine-grained annotations to obtain the state-of-the-arts. 
Thus, more annotations will be better. 
}

\subsection{Limitations}
First, our method currently requires the annotations of position, character string and entity labels of texts in a document, and the labeling process is cost-expensive.
We will resort to semi/weakly-supervised learning algorithms to alleviate the problem in the future.
Another limitation is that the multi-modal context block captures context in the instance granularity, which can be much more fine-grained if introduced token/ character granularity context.
Much more fine-grained context is beneficial to extracting entities across text instances.
}

%% file: sections/6_conclusion.tex
\section{Conclusion} \label{conclusion}
In this paper, we present an end-to-end trainable network integrating  text reading and information extraction for document understanding.
These two tasks can mutually reinforce each other via a multi-modal context block, i.e., 
the multi-modal features, like visual, layout and textual features, can boost the performances of information extraction, while the loss of information extraction can also supervise the optimization of text reading.
On various benchmarks, from structured to unstructured text type and fixed to variable layout, the proposed method significantly outperforms previous methods.
To promote the VRD understanding research, we provide four kinds of benchmarks along the dimensions of layout and text type, and also contribute four groups of strong baselines for the future study.